\pdfoutput=1

\documentclass[11pt]{article}
\usepackage{dsfont}
\usepackage{ACL2023}
\usepackage{enumitem}
\usepackage{times}
\usepackage{booktabs}
\usepackage{latexsym}
\usepackage{graphicx}
\usepackage{diagbox}
\usepackage{float}
\usepackage[table]{xcolor}
\usepackage[namelimits]{amsmath} 
\usepackage[T1]{fontenc}
\usepackage{adjustbox}

\usepackage{subcaption}
\usepackage[utf8]{inputenc}

\usepackage{microtype}
\usepackage{booktabs}
\usepackage{multirow}
\usepackage{multicol}
\usepackage{comment}
\usepackage{amssymb}
\usepackage{inconsolata}

\usepackage{graphicx}
\usepackage{array}

\title{TACO: Enhancing Multimodal In-context Learning via Task Mapping-Guided Sequence Configuration}

\author{Yanshu Li\textsuperscript{1}, Jianjiang Yang\textsuperscript{2}, Tian Yun\textsuperscript{1}, Pinyuan Feng\textsuperscript{3}, Jinfa Huang\textsuperscript{4}, Ruixiang Tang\textsuperscript{5}\thanks{\; Corresponding author.}\\
  \textsuperscript{1}Brown University, \textsuperscript{2}University of Bristol, \textsuperscript{3}Columbia University,\\
 \textsuperscript{4}University of Rochester, \textsuperscript{5}Rutgers University\\
\texttt{yanshu\_li1@brown.edu, ruixiang.tang@rutgers.edu}\\}

\begin{document}
\maketitle
\setlist{noitemsep, topsep=0pt, parsep=0pt, partopsep=0pt}
\begin{abstract}
Multimodal in-context learning (ICL) has emerged as a key mechanism for harnessing the capabilities of large vision–language models (LVLMs). However, its effectiveness remains highly sensitive to the quality of input ICL sequences, particularly for tasks involving complex reasoning or open-ended generation. A major limitation is our limited understanding of how LVLMs actually exploit these sequences during inference. To bridge this gap, we systematically interpret multimodal ICL through the lens of \textbf{task mapping}, which reveals how local and global relationships within and among demonstrations guide model reasoning. Building on this insight, we present \textbf{TACO}, a lightweight transformer-based model equipped with task-aware attention that dynamically configures ICL sequences. By injecting task-mapping signals into the autoregressive decoding process, TACO creates a bidirectional synergy between sequence construction and task reasoning. Experiments on five LVLMs and nine datasets demonstrate that TACO consistently surpasses baselines across diverse ICL tasks. These results position task mapping as a novel and valuable perspective for interpreting and improving multimodal ICL.
\end{abstract}
\begin{figure*}[ht]
    \centering
    \includegraphics[width=1\textwidth]{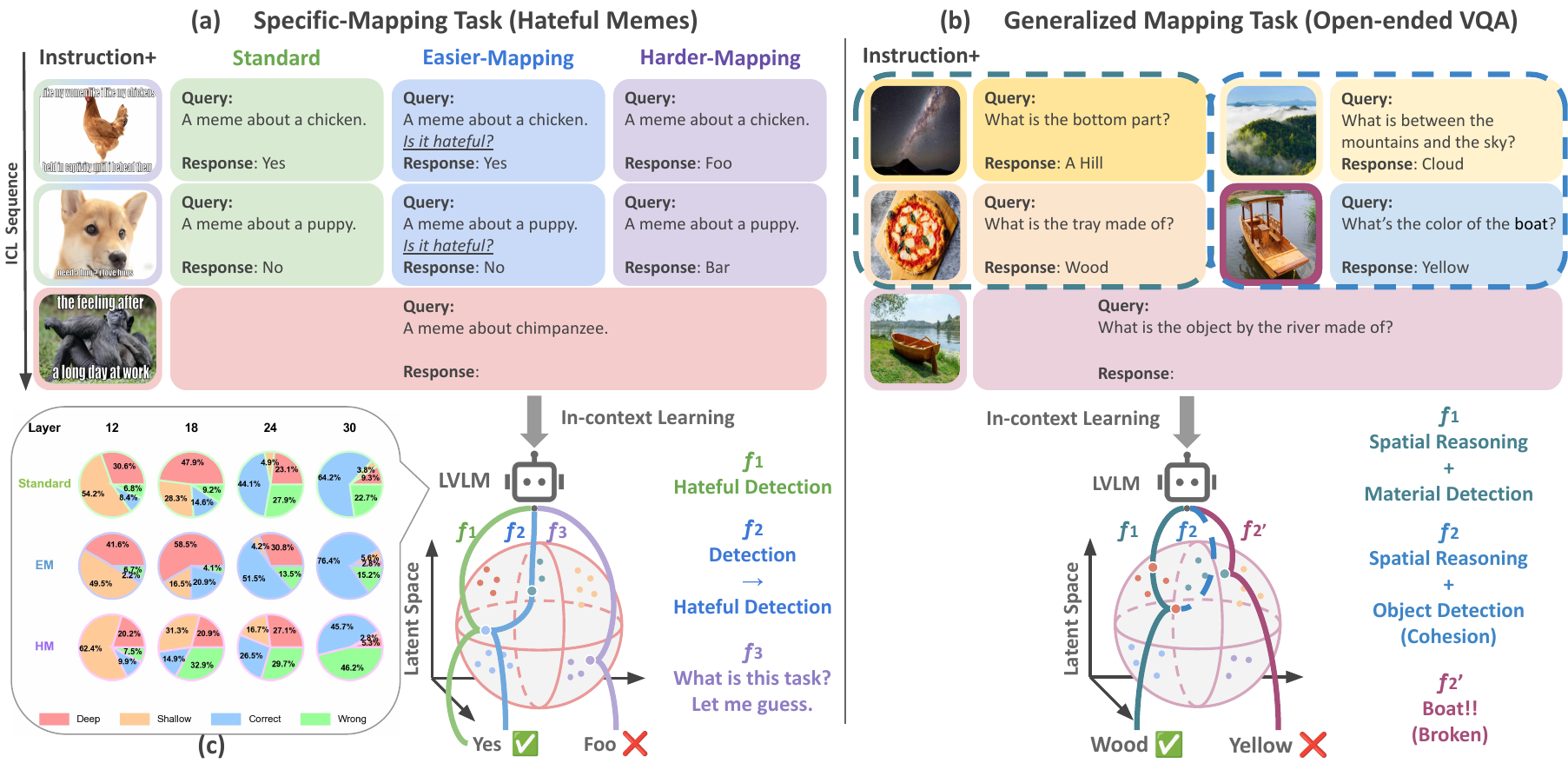}
    \caption{Examples of 2-shot multimodal ICL. (a) In specific-mapping tasks, the ICDs' local mappings are relatively consistent, and the ICL sequence’s global mapping matches them. Their clarity directly affects the LVLM’s reasoning process. The in-context lens in (c) also reflects this latent reasoning shift induced by task mapping. (b) In generalized-mapping tasks, LVLM needs to integrate each local mapping into a cohesive global mapping for reasoning. Overreliance on isolated features (e.g., the visual cue of a boat) can break this cohesion.}
    \label{first}
\end{figure*}

\section{Introduction}

In-context learning (ICL) is a paradigm in which models make predictions by conditioning on a sequence of input–output demonstrations, without updating their parameters. This approach enables models to rapidly adapt to new tasks using only a few examples provided at inference time. Initially, ICL gained significant traction in the domain of large language models (LLMs), where it has demonstrated impressive performance across a wide range of tasks \cite{olsson2022incon, garg2023transformers}. More recently, the concept has been extended to the multimodal setting, giving rise to multimodal ICL, where large vision–language models (LVLMs) learn from interleaved image–text sequences and support complex multi-image reasoning. This capability has become a cornerstone of modern LVLMs \cite{alayrac2022flamingo, intern2.5, qwen2.5}, enabling more flexible and generalizable multimodal understanding.

Despite significant progress in multimodal ICL, configuring effective input sequences remains an open challenge. A standard ICL sequence consists of an instruction, a set of in-context demonstrations (ICDs), and a query sample (see Figure~\ref{first}). Prior studies have shown that even small changes to these sequences can substantially alter LVLM predictions \cite{10350488, zhou2024visual, li2}. These findings highlight the need for robust configuration strategies. However, as multimodal ICL involves diverse cross-modal interactions, our mechanistic understanding is still limited. As a result, existing methods depend on hand-crafted metrics to assess each ICD’s contribution to LVLM reasoning \cite{iter2023in, 10448239}. Rather than relying on specific metrics, we propose a more effective model-centric alternative. Specifically, we address two research questions:

\textbf{How do multimodal sequences influence the ICL performance of LVLMs? (§\ref{MI})}  
We introduce \textbf{task mapping} as a new lens for understanding how input sequences drive multimodal ICL. In the model’s latent space, each ICD defines a \textbf{local} task mapping from its modalities to its output, and these are synthesized into a \textbf{global} task mapping that produces the query response. To investigate this, we develop a probing framework that measures how LVLMs exploit these mappings across sequences. Our study yields two insights: (1) task mapping is essential for effective multimodal ICL, as it guides the alignment between input–output patterns across ICDs and the query; and (2) LVLMs perform better when ICDs form a cohesive task structure, especially in complex cross-modal scenarios. These findings establish task mapping as a principled lens for analyzing and enhancing multimodal ICL.

\textbf{How can we enhance the ICL sequence configuration for effective task mapping? (§\ref{method})}  
Grounded in our theoretical analysis of task mapping, we propose \textbf{TACO} (Task-Aware model for in-COntext Learning), a lightweight transformer-based model that explicitly incorporates task mapping into the selection of ICDs. TACO first encodes the query and instruction to infer task intent, then retrieves ICDs that are both semantically relevant and aligned in reasoning steps. A specialized attention mechanism highlights ICDs that support a coherent interpretation, and layer-wise refinement lets ICDs reinforce one another, producing sequences that enable consistent task reasoning. This task-aware configuration significantly improves the robustness and accuracy of multimodal ICL. Extensive experiments with five advanced LVLMs and nine datasets demonstrate TACO’s superior performance, validating its effectiveness and generality.

Our main contributions can be summarized in three-fold:
\begin{itemize}
    \item To fill the gap in research on multimodal ICL mechanisms in LVLMs, we propose a task mapping framework that systematically captures task distributions across ICDs within an ICL sequence. This framework not only offers a unified view of multimodal ICL under diverse scenarios but also sheds light on the internal behavior of LVLMs during ICL.
    \item Within this task-mapping framework, we propose TACO, a lightweight transformer-based model designed to adaptively retrieve and arrange ICDs from a dataset, yielding optimal ICL sequences for a target LVLM. Evaluations on five LVLMs and nine benchmarks show that TACO achieves superior performance over prior configuration strategies.
    \item We carry out an extensive analysis and ablation study of TACO, isolating the role of each component and design decision. The results provide insights into TACO’s underlying mechanisms and further demonstrate the effectiveness of our task mapping framework for ICL enhancements and applications.
\end{itemize}
\section{Task Mapping in Multimodal ICL}
\label{MI}
In this section, we focus on exploring task mapping in ICL. We first define task mapping (\S \ref{2.1}) and, through systematic empirical experiments, evaluate its impact on multimodal ICL (\S \ref{2.2}) and uncover how LVLMs leverage task mapping across the entire ICL sequence (\S \ref{bas}). All experiments in this section are conducted on two LVLMs: OpenFlamingov2-9B \citep{awadalla2023openflamingo} and Idefics2-8B \citep{ide2}. Results are reported as the average across both models.

\subsection{Identifying Task Mapping in multimodal ICL}
\label{2.1}
\textbf{Notations.} In this work, we mainly focus on ICL for image-to-text tasks, where ICL sequences are organized in an interleaved image-text format. Toward a unified template for various tasks, we reformat ICDs as triplets $(I,Q,R)$, where $I$ is an image, $Q$ is a task-specific text query, and $R$ is the ground-truth response. The query sample is denoted as $(\hat{I},\hat{Q})$. Formally, ICL can be represented as:
\begin{equation}
\adjustbox{width=\columnwidth}{$
\hat{R}\gets \mathcal{M}(S^{n}) =\mathcal{M}(Inst;\underbrace{(I_{1},Q_{1},R_{1}),...,(I_{n},Q_{n},R_{n})}_{n \times ICDs};(\hat{I},\hat{Q})),
$}
\end{equation}
where $\mathcal{M}$ is a pretrained LVLM, $S^{n}$ is an ICL sequence consisting of an instruction $Inst$, $n$-shot ICDs and a query sample, as shown in Figure \ref{first}. 

\textbf{Task Mapping Definition.} We define \textbf{task mapping} as a model-learnable inferential process that transforms input modalities into their outputs within the LVLM’s latent space, capturing both local and global relationships in ICL. Each ICD $(I_{i},Q_{i},R_{i})$ possesses a local task mapping$f$:
\begin{equation}
    f_{i}:(I_{i},Q_{i})\to R_{i},i=1,2,...,n.
\end{equation}
specifying how its image and query jointly map to the response. Then LVLM’s generation on the target query sample $(\hat{I},\hat{Q})$ can be viewed as a global task mapping $\hat{f}$:
\begin{equation} 
\hat{f}:(\hat{I},\hat{Q}) \to \hat{R}. \end{equation} 

Task mapping is inherently indeterminate and complex. To enable systematic analysis, we first consider a scenario where all local mappings $f_{i}$ are nearly identical and coincide with the query’s target mapping. This setup is common for tasks that require the LVLM to follow a fixed reasoning path. We term these as \textbf{specific-mapping tasks}. In such tasks, $I$, $Q$, and $R$ often exhibit structural consistency, which facilitates component-level analysis. 

\textbf{Visualization.} We employ a specific‐mapping task, HatefulMemes \cite{hatefulkiela2020hateful}, to reveal task mapping. Here, each local mapping $f_{i}$ is defined by a binary classification: given a meme image with its caption, determine whether it contains harmful content and output "yes" or "no." (Figure \ref{first}(a)) We use the validation split as our query set and sample $n$ ICDs from the training split using Random Sampling (RS) with a normal distribution to configure the ICL sequences. To highlight $f_{i}$, we create two setups:
    \begin{itemize}[noitemsep, leftmargin=1.5em]
      \item Easier Mapping (\textit{EM}): Augment $Q_{i}$ with an explicit task hint ``Is it hateful?''.
      \item Harder Mapping (\textit{HM}): Replace \(R_i\) (yes/no) with non-semantic words \texttt{foo/bar}.
    \end{itemize}

To explore how task mapping influence LVLM inference, we introduce in-context lens based on logit lens. It defines four anchor word categories: “Shallow” for superficial task recognition, “Deep” for deeper recognition, “Correct” for the query’s true answer, and “Wrong” for its opposite (details in Appendix \ref{log}). Figure \ref{first}(c) visualizes the evolution of internal token outputs under varying task mappings, illustrating the model’s reasoning process. It shows that \textit{EM} greatly enhances the model's ability for deeper task recognition, while \textit{HM} leads to a persistent lack of task awareness, causing the model to rely on random guessing.
\subsection{Task Mapping is Key to Multimodal ICL}
\label{2.2}
To further examine task mapping’s role in multimodal ICL under specific-mapping tasks, we isolate the individual contributions of labels (i.e., “yes” vs. “no”), image features, and task mappings to LVLM performance on HatefulMemes. 

\textbf{Setups.} We introduce targeted ablation settings that selectively impair label reliability and visual clarity, allowing an evaluation of whether LVLMs primarily rely on task mapping over these isolated factors. Specifically, we define: 1. Wrong Labels (\textbf{WL}): Invert 75\% $R_{i}$ labels (yes\(\leftrightarrow\)no). 2. Blurred Images (\textbf{BI}): Applying Gaussian blur to all $I_{i}$. We also apply \textit{EM} solely to $\hat{Q}$, denoted as \textit{EM($\hat{Q}$)}. \textit{BI$(\hat{I})$} refers to applying \textit{BI} solely to $\hat{I}$. The details for these settings are provided in Appendix \ref{setting}. 

\textbf{Results.} Figure \ref{pdis} shows the LVLM’s ICL performance on the same sequences under different settings. The findings are as follows:

\begin{figure}[t]
  \includegraphics[width=1\columnwidth]{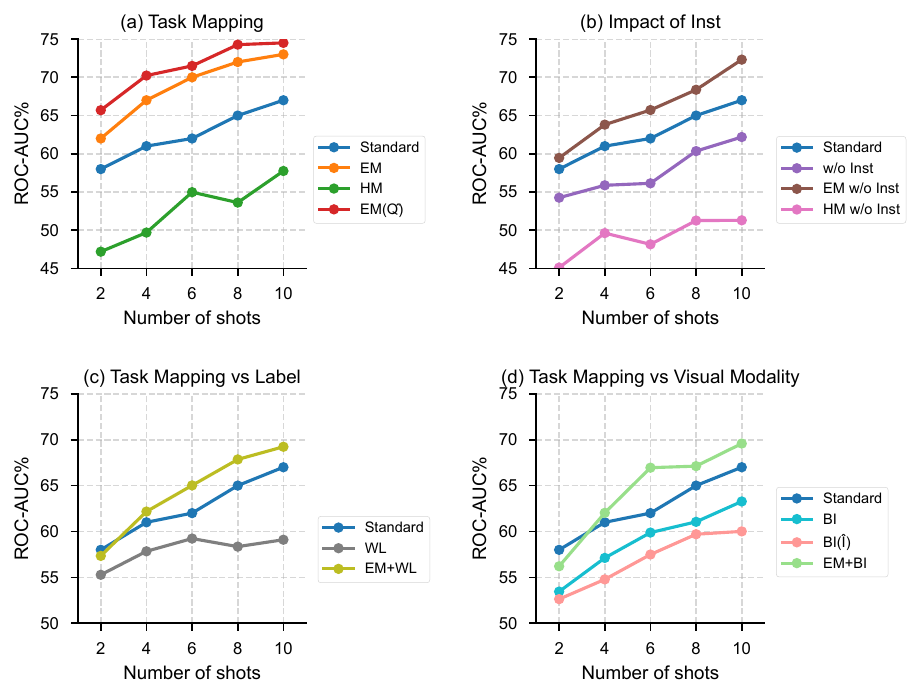}
  \caption{Results on HatefulMemes under various settings. "+" denotes combining two settings.}
  \label{pdis}
\end{figure}

\textbf{Better capturing task mapping consistently improves performance.} As shown in Figure \ref{pdis}(a), across all shot counts, \textit{EM} > \textit{Standard} > \textit{HM} in a clear descending order. This aligns with our observations from in-context lens. In Figure \ref{pdis}(b), removing instructions, which serve as higher-level guidance enabling LVLMs to more deeply capture and utilize $f_{i}$, generally lowers performance. Yet “\textit{EM} w/o $Inst$” still surpasses \textit{Standard}. 

\textbf{Query sample is pivotal.} Surprisingly, Figures \ref{pdis}(a) and (d) show that modifying $\hat{I}$ or $\hat{Q}$ causes greater performance variations than altering all ICDs. We hypothesize that LVLMs prioritize analyzing the query sample and use pretrained knowledge to constrain global task mapping accordingly.

\textbf{Task mapping outweighs labels and visual cues.} In the $\textbf{WL}$ setting, performance drops (Figure \ref{pdis}c), yet stronger task mappings fully recover it. Likewise, in the $\textbf{BI}$ setting, the loss is completely offset by enhanced mappings (Figure \ref{pdis}d). This suggests that both labels and visual modality affect multimodal ICL, but better utilization of task mapping can yield significant performance gains to address deficiencies in unimodal information.

\subsection{ICL Needs Cohesive Global Mapping}
\label{bas}
Building on the central role of task mapping in multimodal ICL, we introduce \textbf{generalized-mapping tasks} to capture real-world challenges in which $f_{i}$ exhibits nuanced or broad variability. Unlike specific mapping tasks, which represent a special case, they involve greater diversity in $Q_{i}$ and $R_{i}$, making component-level manipulation difficult. We therefore adopt a sequence-level analysis, illustrated on the open-ended VQAv2 task \cite{vqav2goyal2017making}.

\begin{figure}[t]
  \centering
  \includegraphics[width=\columnwidth]{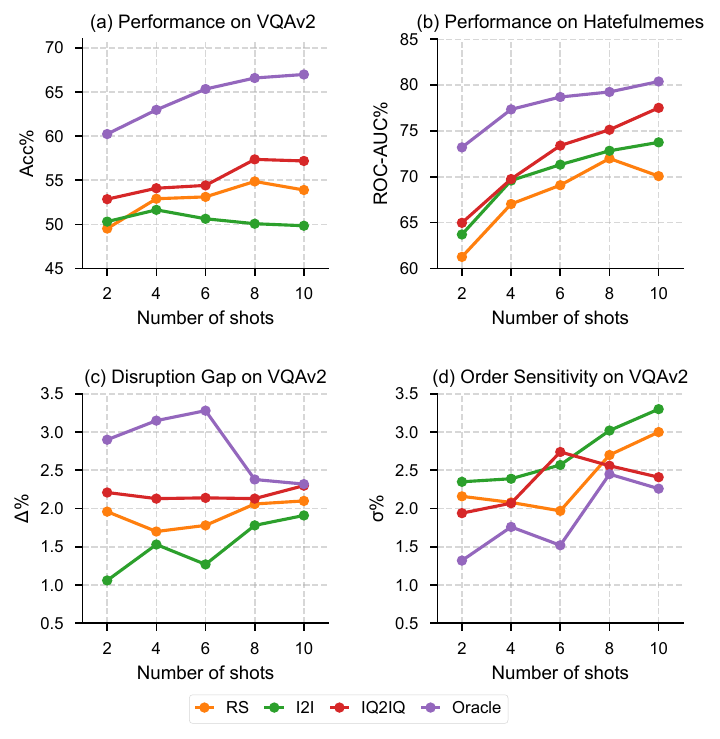}
  \caption{(a-b) Results of different ICL sequence configuration methods on VQAv2 and HatefulMemes. (c-d) Task mapping cohesion analysis of different ICL sequence configuration methods on VQAv2.}
  \label{pdi}
\end{figure}

\textbf{Setups.} Three sequence configuration methods are evaluated: Random Sampling (RS), similarity-based retrieval, and an idealized Oracle. In similarity-based retrieval, ICDs are selected by CLIP cosine similarity, using either image-only alignment (I2I) or joint image query alignment (IQ2IQ). The Oracle method greedily chooses each ICD to maximize the log likelihood of generating the ground-truth response while accounting for the cumulative influence of prior demonstrations (computational details in Appendix \ref{self}). 

\textbf{Hypothesis.} Figure \ref{pdi}(a) and (b) show that multimodal alignment via IQ2IQ consistently outperforms unimodal alignment (I2I) and RS on both datasets. Meanwhile, Oracle consistently achieves the highest accuracy. An unexpected finding is that I2I performs worse than RS on VQAv2 but not on HatefulMemes. We hypothesize that \textbf{task mapping cohesion} explains this phenomenon. In generalized-mapping tasks, effective ICL requires ICDs to jointly support complex reasoning. When performing ICL with certain sequences configured via I2I, isolated feature matching introduces a fragmented reasoning bias that leaves the sequence’s global mapping cohesion \textbf{broken}.

\textbf{Proof.} To validate this hypothesis, we introduce two metrics for measuring task mapping cohesion: Disruption Gap (\textbf{$\Delta$}) and Order Sensitivity (\textbf{$\sigma$}) (details in Appendix \ref{sensitive}). These metrics reflect the impact of cohesive task mapping on multimodal ICL, with higher $\Delta$ and lower $\sigma$ indicating stronger reliance on cohesive task mapping. Figure \ref{pdi}(c-d) shows that Oracle achieves the highest $\Delta$ and lowest $\sigma$ across all shots, proving its ability to construct cohesive sequences through holistic consideration of preceding ICDs. However, as shots increase to 8 and 10, Oracle's $\Delta$ surges while $\sigma$ plunges, revealing potential local optimization issues and accumulated bias in longer sequences. Meanwhile, I2I consistently underperforms RS on both metrics, while IQ2IQ surpasses RS but remains unstable, aligning with accuracy trends in generalized-mapping tasks and supporting our hypothesis.

Finally, based on performance, $\Delta$ and $\sigma$, we can categorize all experimental ICL sequences into four types for case studies (see Appendix \ref{case}): (1)-(2) sequences impaired by isolated dependencies (e.g., similar image features and local mapping bias), (3) sequences resembling specific-mapping tasks, and (4) the most common type, featuring diverse local mappings that collectively enhance cohesive task mapping. Such diversity enables LVLMs to overcome excessive reliance on superficial features and achieve superior multimodal ICL performance.

\begin{figure*}
    \centering
    \includegraphics[width=1\textwidth]{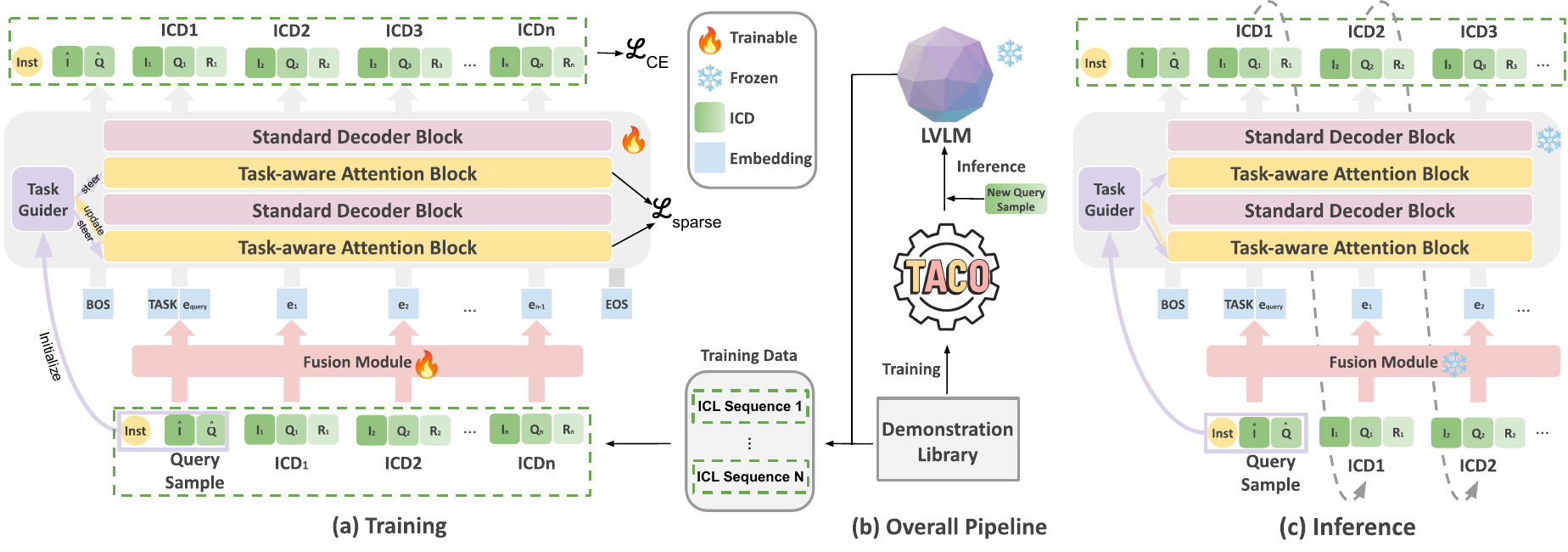}
    \caption{Our overall pipeline, shown in (b), consists of three parts: a demonstration library, TACO, and a pre-trained LVLM. TACO treats each $(I,Q,R)$ example in the demonstration library as a token. (a) shows TACO training using the LVLM-constructed training data. (c) shows that, given a new query sample, TACO autoregressively retrieves samples from the demonstration library to form a high-quality ICL sequence for LVLM inference.}
    \label{pipe}
    \vspace{-10pt}
\end{figure*}

\section{The Proposed Method: TACO}
\label{method}
\textbf{Motivation.} Task mapping plays a crucial role in enabling effective ICL for LVLMs, as discussed in §\ref{MI}. Thus, to construct high-quality ICL sequences, two objectives must be met: (1) each ICD should contribute a meaningful local mapping that supports the reasoning process, and (2) the sequence as a whole should form a cohesive global mapping that aligns with the target task. These objectives reflect how models implicitly organize and utilize contextual information during inference. However, existing metric-centric methods may not fully model these mappings, as shown by the results in §\ref{bas}. Oracle that directly leverages the LVLM’s own inference capability consistently outperforms similarity-based methods. Although Oracle's reliance on the ground-truth response makes it impractical for direct inference, it provides an effective way to generate training data for model-centric learning of LVLM reasoning paths. Therefore, we propose the \textbf{Task-Aware model for in-COntext Learning (TACO)}, a lightweight, end-to-end framework designed to select ICDs that enhance both local and global task alignment. TACO is trained using data derived from LVLMs and leverages a specialized attention mechanism that models the reasoning patterns guiding task mapping. 

Figure~\ref{pipe} illustrates the overall architecture of TACO. Its backbone consists of four transformer decoder blocks. Each triplet example $(I_i, Q_i, R_i)$ from the demonstration library $DL$ is treated as a distinct token. The training dataset $D_S$ is composed of $N$-shot ICL sequences. During inference, given a query sample and $Inst$, TACO can autoregressively retrieve $n$ samples from $DL$ to configure the optimal $n$-shot ICL sequence.

\textbf{Input Embedding.} Let $x_{i}$ denote $i$-th ICD token $(I_{i}, Q_{i},R_{i})$ and $\hat{x}$ denote the query sample $(\hat{I},\hat{Q})$. In each input sequence, $\hat{x}$ is placed ahead of all $x_{i}$. To align with the autoregressive generation process, we use two special tokens, $[BOS]$ and $[EOS]$, to mark the beginning and end of the input sequence during training. These tokens are added to TACO's vocabulary. We also introduce a $[TASK]$ token into the vocabulary and concatenate it with $\hat{x}$ in the input sequence. It acts as a semantic anchor for task mapping. Therefore, for a given $S^{N}$, we reconstruct it as a token sequence $([BOS], [TASK] + \hat{x}, x_{1}, ..., x_{N}, [EOS]\})$. 
To filter and balance multimodal features for better mapping construction, we employ a binary fusion module to generate the embedding $e_{i}$ for $x_{i}$:
\begin{equation}
f_{i}=\sigma (W_{f} \cdot [E_{I}(I_{i}) \oplus E_{T}(Q_{i}\oplus R_{i})]+b_{f}),
\end{equation}
\begin{equation}
e_{i} = f_{i} \cdot E_{I}(I_{i})+ (1-f_{i}) \cdot E_{T}(Q_{i}\oplus R_{i}),
\end{equation}
where $E_{I}(\cdot)$ and $E_{T}(\cdot)$ denote image encoder and text encoder of CLIP. Finally, the input embedding sequence of TACO is presented as follows:
\begin{equation}
e_{S^{N}} = [e_{\text{BOS}}, \hat{e}, e_{1}, \dots, e_{N}, e_{\text{EOS}}],
\end{equation}
where $e_{\text{BOS}}$ and $e_{\text{EOS}}$ are learnable embeddings of $[BOS]$ and $[EOS]$. $\hat{e}$ is a joint representation formed by concatenating the learnable embedding of $[TASK]$ with the embedding of $\hat{x}$ generated using the same fusion module. The index of $\hat{e}$ is always 1, and $I_{idx}$ denotes the index set of $e_{i}$.

\textbf{Task-aware Attention.} The task-aware attention in TACO enables dynamic ICL sequence configuration by integrating task mappings into attention computation. Its core is the task guider ($TG$), an embedding independent of the input sequence, designed to capture fine-grained global task mapping within ICL sequences. $TG$ encodes task intent through initialization by the multimodal fusion of the query sample and instruction:
\begin{equation}
e_{TG}^{(0)}=W_{TG}\cdot(E_{I}(\hat{I})\oplus E_{T}(\hat{Q}) \oplus E_{T}(Inst')),
\end{equation}
where $W_{TG} \in \mathbb{R}^{d \times 3d}$ is a learnable weight matrix used to regulate the entire $TG$. $Inst'$ is a simplified instruction generated by GPT-4o (Appendix \ref{instapp}). 

Task-aware attention is applied selectively to certain layers, denoted as $L_{Ta}$. At each of these layers, $TG$ steers the attention mechanism by weighting relevance scores, which are derived from the interaction between $TG$ and token embeddings. This interaction captures the hierarchical relationships between task mappings within the ICL sequence:
\begin{equation}
    t_i^{(l)} = \sigma\Bigl(\mathrm{MLP}^{(l)}\bigl(e_{TG}^{(l)} \oplus e_i\bigr)\Bigr),
\end{equation}
where $\mathrm{MLP}^{(l)}$: $\mathbb{R}^{2d} \rightarrow \mathbb{R}^{d}$ is a layer-specific network producing a scalar weight $t_{i}^{(l)} \in [0,1]$ and $\sigma$ is the sigmoid function. This weight reflects the degree to which each token contributes to the cohesive task mapping, dynamically adapting TACO's attention to emphasize semantically salient features. It modulates attention logits through a task-aware mask $M^{(l)}$. For intra-ICD tokens, the mask scales pairwise cosine similarities by $-\log(t_{i}^{(l)})$. For query-ICD tokens, a learnable coefficient $\alpha$ allows $\hat{e}$ to guide attention throughout the sequence. The mask is computed as follows for position $(i,j)$:
\begin{equation}
\adjustbox{width=\columnwidth}{$
M_{ij}^{(l)} =
\begin{cases}
\frac{\mathrm{sim}(e_i,\, e_j)}{\sqrt{d}} \;\cdot\;
\bigl(-\log(t_i^{(l)})\bigr),
&  j \le i \text{ and } i,j \in I_{idx}, \\[4pt]
\frac{\alpha \mathrm{sim}(\hat{e},\, e_j)}{\sqrt{d}} \;\cdot\;
\bigl(-\log(t_i^{(l)}) \bigr),
&  i = 1 \text{ and } j \in I_{idx},\\[4pt]
-\infty, 
& \text{otherwise}.
\end{cases}$}
\end{equation}
Here, the first case emphasizes interactions between local task mappings, and the second case enables deep task mapping cohesion. The last case preserves the autoregressive nature.
The mask $M^{(l)}$ is integrated into standard attention, forming task-aware attention (TaAttn), as follows:
\begin{equation}
\text{TaAttn}(Q, K, V) = \text{softmax} \left( \frac{QK^T}{\sqrt{d}} + M^{(l)} \right) V.
\end{equation}

In particular, $TG$ is updated between task-aware layers to preserve task mapping, enabling hierarchical refinement from coarse task intent to fine-grained mapping. After processing layer $l \in L_{Ta}$ through residual connections, $TG$ is updated via:
\begin{equation}
e_{TG}^{(l')} = \operatorname{LN} \left( e_{TG}^{(l)} + \operatorname{Attention} (e_{TG}^{(l)}, H^{(l)}) \right),
\end{equation}
where $l'$ denotes the next task-aware layer in $L_{Ta}$, $H^{(l)}$ denotes the hidden states of layer $l$ and $\operatorname{LN}$ denotes layer normalization. $\operatorname{Attention}(\cdot,\cdot)$ denotes the cross-attention computation. To ensure focused attention patterns, we introduce a sparsity loss that penalizes diffuse distributions:
\begin{equation}
\mathcal{L}_{\text{sparse}} = \sum_{l \in L_{Ta}} \frac{1}{N} \sum_{i=1}^{N} \text{KL} \left( \text{softmax}(M_{i:}^{(l)}) \parallel \mathcal{U} \right),
\end{equation}
where $\mathcal{U}$ is a uniform distribution. Minimizing this KL divergence prompts a sharper representation of task-mapping. The total training objective combines the standard cross-entropy loss for sequence generation, sparsity regularization, and L2-norm constraint on $TG$ to prevent overfitting:
\begin{equation}
\mathcal{L} = \mathcal{L}_{CE}+\lambda_{1} \mathcal{L}_{\text{sparse}}+\lambda_{2} \left \| W_{TG}\right \| _{2}^{2}.
\end{equation}

\begin{table*}
\centering
\resizebox{0.9\textwidth}{!}{\begin{tabular}{ c c c c | c c | c | c | c | c}
\toprule
\multirow{3}*{\textbf{Methods}} & \multicolumn{3}{c|}{\textbf{VQA}} & \multicolumn{2}{c|}{\textbf{Captioning}} & \multicolumn{1}{c|}{\textbf{Classification}} & \multirow{2}*{\textbf{Hybrid}} & \multirow{2}*{\textbf{Fast}} & \multirow{2}*{\textbf{CLEVR}}\\
\cmidrule(lr){2-7}
& VQAv2 & VizWiz & OK-VQA &  Flickr30K & MSCOCO  & HatefulMemes &\multirow{2}*{ACC.↑} & \multirow{2}*{ACC.↑}& \multirow{2}*{ACC.↑}\\
& ACC.↑ & ACC.↑ & ACC.↑ &  CIDEr↑ & CIDEr↑  & ROC-AUC↑ & & &\\
\midrule
RS         & 60.32 & 43.38 & 51.85 & 93.67  & 110.81 & 74.14 & 17.74 & 64.04 & 43.42 \\
I2I        & 58.29 & 43.10 & 51.54 & 96.02  & 110.77 & 76.00 & 14.69 & 66.69 & 41.56 \\
IQ2IQ      & 61.37 & 44.96 & 53.90 & 94.24  & 112.04 & 74.83 & 34.85 & \underline{68.26} & 41.58 \\
IQPR       & 62.25 & 44.99 & 54.58 & 95.19  & 113.52 & 73.62 & 35.63 & 67.68 & 43.98 \\
DEmO       & 61.10 & 45.41 & 55.28 & 95.21  & 113.24 & 74.02 & 34.56 & 67.20 & 42.12 \\
Lever-LM   & \underline{64.13} & \underline{48.13} & \underline{58.33} 
           & \underline{98.24} & \underline{118.27} & \underline{78.86} 
           & \underline{41.61} & 67.36 & \underline{46.51} \\
\rowcolor[HTML]{F0F0F0}
Ours       & \textbf{66.75} & \textbf{52.07} & \textbf{61.54} 
           & \textbf{99.62} & \textbf{119.47} & \textbf{80.59} 
           & \textbf{45.22} & \textbf{68.73} & \textbf{48.45} \\
\bottomrule
\end{tabular}}

\caption{\label{main}
Results of different ICL sequence configuration methods across 9 datasets, with both training and generated sequences being 4-shot. Each result is the average performance across \textbf{five} LVLMs with the same prompt format. The highest scores are highlighted in \textbf{bold}.
\underline{Underlined} values indicate the results of the best baselines. Detailed results for each LVLM can be found in Table \ref{detailed}.
}
\vspace{-10pt}
\end{table*}

\textbf{Inference and Prompt Construction.} After training, TACO can autoregressively select demonstrations from a library and configure ICL sequences. Given a new query sample $\hat{x}$, the input sequence to TACO during inference is $\{[BOS], [TASK] + \hat{x}\}$, where $\hat{x}$ is embedded using the trained fusion module. The shot of the generated sequence, denoted as $n$, is a user-defined value. It may differ from the shot count $N$ in $D_{S}$, as discussed in §\ref{shot}. TACO then selects $n$ ICDs using a beam search strategy with a beam size of 3, producing the optimal $n$-shot ICL sequence $S^{n}$. This sequence is used to construct a prompt for LVLMs, formatted as: $\{Inst; ICD_{1}, ..., ICD_{n}; Query \,Sample\}$, which is then used to perform multimodal ICL. Example prompts are provided in Appendix \ref{prompt}.

\section{Experiment}
\subsection{Training Data Construction and Models}
Following standard multimodal ICL evaluation practices \cite{awadalla2023openflamingo}, we select six high-quality datasets across three core VL tasks: VQAv2, VizWiz \cite{vizwizgurari2018vizwiz}, and OK-VQA \cite{okvqamarino2019ok} for open-ended VQA; Flickr30K \cite{flickr30kyoung2014image} and MSCOCO \cite{mscocolin2014microsoft} for captioning; and HatefulMemes for classification. To further assess TACO’s abilities in generalized-mapping tasks, we create a mixed-task dataset \textbf{Hybrid}, by sampling 5,000 instances from the training set from each above dataset, with validation samples drawn proportionally from their validation sets. We also adopt two challenging image-to-text tasks from the latest multimodal ICL benchmark, VL-ICL \cite{zong2024vlicl}: Fast Open-Ended MiniImageNet\footnote{We apply fine-grained adjustments to \textbf{Fast} to increase its difficulty; see Appendix \ref{dL} for details.} (\textbf{Fast}) and \textbf{CLEVR}. 

To construct the high-quality sequence dataset $D_{S}$ for TACO training from the above datasets, we first reformulate them into $(I, Q, R)$ triplets. Using clustering, we select $K$ samples from their training sets as query samples, forming the query set $\hat{D}$. For each query sample in $\hat{D}$, $N$ ICDs are retrieved from the remaining data using the Oracle method described in §\ref{bas}, creating $S^{N}$. This retrieval process is further refined through beam search to improve the quality and diversity of $D_{S}$. The implementation details are provided in Appendix \ref{training}. All $S^{N}$ begin with a CoT-style $Inst$, as detailed in \textit{Beginning1} of Table \ref{instde}.

Our experiments evaluate four advanced open-source LVLMs: OpenFlamingov2-9B, Idefics2-8B, InternVL2.5-8B \cite{intern2.5}, and Qwen2.5VL-7B \cite{qwen2.5}, as well as a closed-source model, GPT-4V \cite{openai2024gpt4v}. Detailed descriptions of the datasets and LVLMs are provided in Appendix \ref{dL}.
\subsection{Baselines and Implementation Details}

We adopt RS and two similarity-based retrieval methods introduced in §\ref{bas} as baselines,  along with three previous SOTA configuration methods: IQPR \cite{sqprli2024configure}, DEmO \cite{demo}, and Lever-LM \cite{yang2024lever}. Lever-LM is a tiny language model with several standard decoder blocks that performs automatic $S^{n}$ configuration. As it also requires model training, we treat it as the primary baseline. For a fair comparison, we set Lever-LM’s depth to four layers. Details of the baselines are provided in Appendix \ref{baseline}.
 
We evaluate ICL sequences on LVLMs using validation sets of the datasets, with the training sequence shot $N$ and the generated sequence shot $n$ set to 4. Query set $\hat{D}$ sizes vary by dataset (Table \ref{size}). We utilize the image and text encoders from CLIP-ViT-L/14 to generate image and text embeddings. For all tasks, we employ a unified encoder training strategy: updating only the last three layers while keeping all preceding layers frozen. TACO training employs a cosine annealed warm restart learning scheduler, AdamW optimizer, 1e-4 learning rate, batch size 128, and runs for 20 epochs.

\begin{table*}
\centering
\resizebox{0.9\textwidth}{!}{\begin{tabular}{lccc|cc|c|c|c|c}
\toprule
\multirow{2}{*}{\textbf{Configuration}} 
& \multicolumn{3}{c|}{\textbf{VQA}} 
& \multicolumn{2}{c|}{\textbf{Captioning}} 
& \textbf{Classification} 
& \multirow{2}{*}{\textbf{Hybrid}}
& \multirow{2}{*}{\textbf{Fast}}
& \multirow{2}{*}{\textbf{CLEVR}} \\
\cmidrule(lr){2-7}
& VQAv2 & VizWiz & OK-VQA
& Flickr30K & MSCOCO
& HatefulMemes 
& &  & \\
\midrule
\rowcolor[HTML]{F0F0F0}
\textbf{Full TACO}       &   \textbf{66.75}    &  \textbf{52.07}     &     \textbf{61.54}  &   \textbf{99.62}    &    \textbf{119.47}   &   \textbf{80.59}    &    \textbf{45.22}   &  \textbf{68.73}     &   \textbf{48.45}    \\
\midrule
(a) w/o [TASK] token   &   64.58    &   50.24    &  60.26     &  98.39   &   118.04    &   79.51   &   42.83    &   67.18   &    46.38  \\
(b) w/o $TG$ updates   &    63.53   & 48.36  & 59.01 &   97.65  &   117.84    &   77.24    &  40.56     &   65.27    &   44.79   \\
(c) w/o $\mathcal{L}_{\text{sparse}}$  
&  63.79  &  51.25   &  60.95  &  98.19  &  118.10 &  78.29 &  42.33  &  65.93   &  45.81 \\    
(d) w/o $\left \| W_{TG} \right \| _{2}^{2}$   &    62.71   & 48.29  & 58.41 &  98.45  &   117.72    &   76.35    &  38.40  &   66.26    &    44.28  \\
\midrule
(e) Random initialization &   59.46    &   42.97    &   54.59    &   94.67    &   111.52    &   74.48   &   34.29   &    60.38  &  42.52  \\
(f) w/o $\hat{I}$  &   64.10    &  49.61   &   59.65    &   97.19    &   115.28    &   78.50    &  41.54     &   66.45    &  45.08   \\
(g) w/o $\hat{Q}$   &   62.54    &  47.24   &   59.47   &   96.95    &   114.73    &   76.83    &  40.22     &    66.07   &   44.93 \\
(h) w/o $Inst'$   &   62.68  &  48.02  &  60.08   &   98.32   &   117.90   &   77.32  &  40.75  & 66.73   &  45.16\\

\bottomrule
\end{tabular}}
\caption{\label{atable}Results of the ablation study on task mapping augmentation of TACO. Specifically, (a)-(d) correspond to diverse task-aware attention construction, (e)-(h) to diverse $TG$ initialization.}
\end{table*}

\begin{figure*}
    \centering
    \includegraphics[width=1\textwidth]{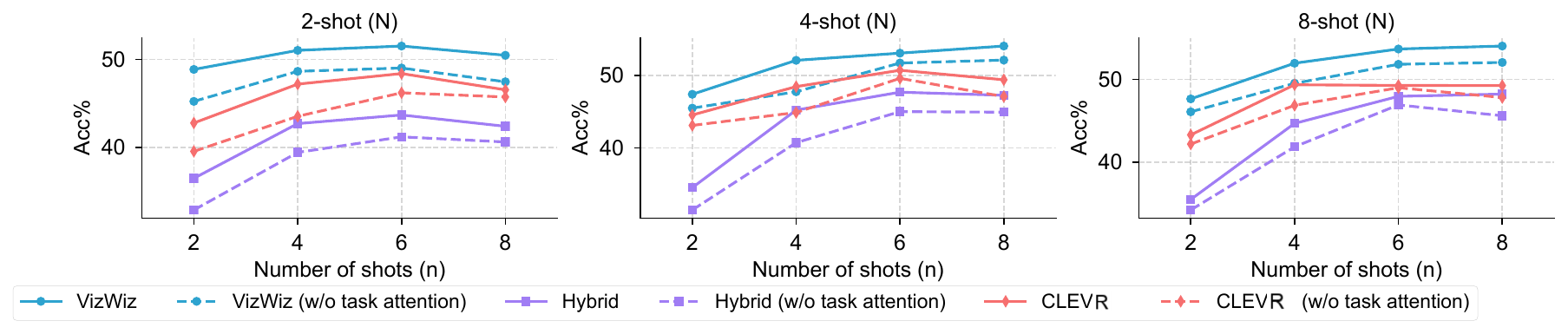}
    \vspace{-15pt}
    \caption{Results of TACO with and without task-aware attention under different $N$-$n$ settings across three datasets, where $N$ is the training sequence shot and $n$ is the generation sequence shot.}
    \label{shots}
    \vspace{-10pt}
\end{figure*}
\subsection{Main Results}
Table \ref{main} summarizes the average performance of ICL in five LVLMs using different methods of configuring the ICL sequence. TACO consistently outperforms all baselines across all nine datasets, highlighting its robustness and effectiveness in fully leveraging the potential of LVLMs for diverse multimodal ICL scenarios. Notably, TACO delivers particularly strong results in generalized-mapping tasks, achieving an average improvement of 3.26\% in VQA tasks, with the second highest gain of 3.61\% observed on \textbf{Hybrid}. These results demonstrate that strengthening task mapping enhances the autoregressive generation process of language models, equipping them with a broader understanding and enabling the construction of more precise cohesive task mappings. In Appendix \ref{mainapp}, we present more evaluations of how ICL sequence configuration affects LVLM using per-model data and include efficiency analyses of TACO to show its low computational overhead.

\section{Ablation Study and Discussions}
\label{shot}
In this section, we examine the impact of task-aware attention and reveal, from a task-mapping perspective, how it enhances ICL performance.

Table \ref{atable} shows that each ablated component induces a complete performance degradation. $TG$, initialized by fusing the query’s bimodal context with instruction semantics, establishes a task intent that aligns with the observation of §\ref{MI}: global mapping synthesis relies on query-driven grounding. Jointly anchored by the $[TASK]$ token, this intent prevents local mapping drift during autoregressive generation but also enables dynamic refinement through layered attention updates. By iteratively resolving coarse task boundaries into fine-grained patterns, $TG$ harmonizes intra-sequence dependencies and query-context interactions, forming a feedback loop where each retrieved ICD sharpens global mapping cohesion. In conclusion, task-aware attention effectively encodes task mapping as a dynamic attention-driven process, transcending static ICD aggregation to achieve consistent performance improvements in multimodal ICL. 

\begin{table*}[t]
  \centering
  \begin{tabular}{lccccccc}
    \toprule
    Method            & VQAv2  & VizWiz & OK-VQA & Flickr30K & MSCOCO  & HatefulMemes & Hybrid  \\
    \midrule
    TACO (RS)        & 62.38  & 47.69  & 54.47  & 98.31    & 115.83  & 76.49        & 38.62   \\
    TACO (I2I)       & 61.95  & 47.28 & 53.86  & 98.74     & 117.25  & 77.46        & 36.90   \\
    TACO (IQ2IQ)     & 64.37  & 50.18 & 59.23  & 99.17     & 118.68  & 78.93    & 41.05   \\
    TACO (Oracle)    & \textbf{66.75}  & \textbf{52.07}  & \textbf{61.54}  & \textbf{99.62}     & \textbf{119.47}  & \textbf{80.59}        & \textbf{45.22}   \\
    \bottomrule
  \end{tabular}
\caption{Results of TACO under diverse training sequence construction strategies.}
  \label{app:data}
\end{table*}

To gain a deeper understanding of the role of task mapping throughout from training to inference, we explore different combinations of training and generation shots. Our findings are as follows:

\textbf{Task mapping consistently enhances multimodal ICL}. Figure \ref{shots} shows that across all $N$-$n$ combinations, task-aware attention always improves performance, highlighting the value of focusing ICL sequences on task mapping.

\textbf{Cohesion remains robust as shots increase}. For specific-mapping tasks (e.g., \textbf{CLEVR}), when $N$ is fixed, performance gains diminish as $n$ increases, while generalized-mapping tasks generally maintain steady improvements. This arises from each new ICD's unique contribution to the global task mapping, potentially deepening it rather than yielding diminishing returns on a specific mapping.

\textbf{Task mapping enables flexibility in $N$ and $n$.} Although task-aware attention works best when $N$ equals $n$, the cohesive design of task mapping allows TACO to effectively interpolate and extrapolate sequence shots across a flexible range of values. This adaptability ensures performance across diverse training data and enhances the model's potential for practical multimodal ICL applications, where flexibility and scalability are critical. 

Next, we investigate the construction of training sequences for TACO. In the main experiments, we use Oracle that approximates an LVLM’s optimal multimodal ICL. Training TACO on sequences generated by Oracle enables it to learn how to configure ICL inputs with correct task mappings, especially at the global level across the demonstration set. To test the robustness of this learning process, we replace Oracle with similarity-based methods and examine how this change affects TACO’s accuracy in inferring task mappings.

As TACO is training-based, one of the most important aspects is high-quality data. Table \ref{app:data} demonstrates that using Oracle as the method to construct the training data is optimal. Since our approach leverages TACO to capture how LVLMs understand and utilize task mapping, the training data that best reflects the internal mechanisms of the LVLM is most effective. Moreover, it can be observed that TACO, when combined with RS, I2I, and IQ2IQ, brings significant performance improvements over using RS, I2I, and IQ2IQ in isolation. This indicates that our method can mitigate the inherent limitations of retrieval strategies through training, further enhancing the practical value of TACO.

In Appendix \ref{ablation}, we conduct additional ablation studies on the construction of input embeddings as well as the format and position of $Inst$. We also evaluate TACO’s extension to NLP and text-to-image tasks. Furthermore, we revisit the theoretical framework introduced in §\ref{MI} through TACO, further confirming its robustness. Together, these experiments demonstrate that the improvements that TACO brings to multimodal ICL are derived from its task-mapping-guided configuration.

\section{Related Works}
\textbf{Interpreting ICL.} The mechanisms of ICL are crucial to better employing it \cite{gao2021making, dong2024survey, li1}. \citet{min2022rethink} attribute ICL's success to explicit information in ICDs like label space and input distribution, while \citet{zhou2023least} emphasize the importance of input-output mappings. To find a unified solution, \citet{wei2023larger} and \citet{pan-etal-2023-context} disentangle ICL into Task Recognition and Task Learning. \citet{zhao2024unveiling} further propose a two-dimensional coordinate system to explain ICL behavior via two orthogonal variables: similarity in ICDs and LLMs' ability to recognize tasks. However, these studies are often confined to specific-mapping tasks with small label spaces and struggle to address complex multimodal scenarios. We present related work on configuring ICL sequences in Appendix \ref{app:rela}.

\section{Conclusion}
In this work, we systematically demonstrate the principles and critical role of task mapping within ICL sequences for enabling effective multimodal ICL in LVLMs. These insights further motivate the use of task mapping to explore more effective ICL sequence configuration strategies that truly align model learning behavior and internal demands. To this end, we propose a transformer-based model, TACO, which employs task-aware attention to deeply integrate task mapping into the autoregressive process, thereby optimizing sequence configuration. Experiments show consistent outperformance over SOTA baselines, particularly in generalized-mapping tasks. This study not only presents a practical model but also provides the multimodal ICL community with a new and reliable research direction.

\section*{Limitations}
\paragraph{Perspectives from cognitive science are not yet incorporated.} In this work we propose task mapping, which represents an abstract inference process performed by an LVLM in its latent space. This concept aligns with themes in cognitive science. By thoroughly examining task mapping we may discover ways to equip LVLMs with more advanced cognitive capabilities. However, our current study does not incorporate cognitive science theory or pursue interdisciplinary exploration, which somewhat limits the impact of task mapping. In future research we will explore cross disciplinary integration based on task mapping.

\paragraph{Our analysis does not examine the internal mechanisms through which task mapping is realized.} This study does not delve into the role of LVLMs' internal attention mechanisms and hidden state in capturing and utilizing task mapping. Investigating how task mapping manifests within attention layers could uncover deeper connections between sequence configuration and model reasoning, offering another promising avenue for our future work.
\section*{Acknowledgements}
We extend our sincere gratitude to Prof. Ellie Pavlick from Brown University for her constructive suggestions on the empirical design of this study. Her expertise and insights were invaluable to this research.
\bibliography{acl2023}
\bibliographystyle{acl_natbib}

\newpage
\appendix

\section{Additional Related Works}
\label{app:rela}
\textbf{Configuring ICL sequences.} To configure high-quality ICL sequences that bolster multimodal ICL in LVLMs, researchers have explored numerous methods, with metric-centric approaches emerging as the most prominent \cite{order}. The most direct metric for both implementation and evaluation is similarity. In this category, methods select ICDs from a demonstration library by comparing their embeddings, an approach widely used in retrieval augmented generation (RAG) systems \cite{luo2024inde, chen2024learning}. Retrieval strategies based on semantic entropy have also been applied to tasks demanding more fine-grained selection criteria \cite{wu2023self, entro}. To accommodate more complex tasks, several novel metrics have been introduced. \citet{demo} define an influence score, which quantifies the change in model confidence induced by each demonstration, and combine this score with entropy to configure ICL sequences. \citet{opti} leverages log probabilities of LLM-generated outputs to systematically prune the search space of possible orderings. Although these human-designed metrics can partially capture each ICD’s contribution to latent reasoning, the black-box nature of LVLMs prevents them from fully reflecting the model’s internal inference. For example, similarity-based methods may not provide LLMs with deep task mappings \cite{liu2021makes, sqprli2024configure}. Approaches designed to mitigate ICD bias can also inadvertently introduce new biases \cite{lyu2023zicl, yuan2024llm}. Model-centric methods have also emerged later, employing multiple models for more demanding selection \cite{wu2023self, wang2024learning, s2024incon}. These methods are not end-to-end and overly focus on ICD selection over ordering. One work closely connected to ours is \citet{yang2024lever}, which introduces a tiny language model composed of two encoder blocks to automatically select and order ICDs. However, its effectiveness on complex tasks is constrained by a lack of deep insight into task mapping. 

\section{Formal Theoretical Definition}
In \S \ref{2.1}, we provide simple definitions of local and global task mappings for clarity. Here, we develop a more complete theoretical analysis of task mapping.

Let \(\mathcal I\) denote the image space, \(\mathcal Q\) the query space, and \(\mathcal R\) the response space.  Define the space of deterministic mappings
\[
\mathcal F \;=\;\bigl\{\,f:\mathcal I\times\mathcal Q\to\mathcal R\,\bigr\}.
\]

The pretrained LVLM \(M_\theta\) induces a conditional distribution over \(\mathcal R\) given an \(n\)-shot ICL prompt:
\begin{equation}
\resizebox{\columnwidth}{!}{$
p_\theta(r \mid S^n)
=\Pr_{M_\theta}\bigl(r \mid \mathrm{Inst};D_1,\dots,D_n;(\hat I,\hat Q)\bigr),
$}
\label{eq:dist}
\end{equation}
where 
\(\;S^n=(\mathrm{Inst};D_1,\dots,D_n;(\hat I,\hat Q))\) 
and \(D_i=(I_i,Q_i,R_i)\).

\medskip
\noindent\textbf{Definition B.1 (Local Task Mapping).}  
Each demonstration \(D_i\) induces a local mapping \(f_i\in\mathcal F\) defined by
\begin{equation}
\resizebox{\columnwidth}{!}{$
f_i
=\arg\max_{f\in\mathcal F}
\;\mathbb E_{(I,Q,r)\sim D_i}\bigl[\mathds{1}\{f(I,Q)=r\}\bigr]
\quad\Longrightarrow\quad
f_i(I_i,Q_i)=R_i,
$}
\label{eq:local}
\end{equation}
which under \(M_\theta\) equivalently satisfies
\begin{equation}
\resizebox{\columnwidth}{!}{$
f_i(I,Q)
=\arg\max_{r\in\mathcal R}\;
p_\theta\bigl(r\mid \mathrm{Inst};D_{<i},(I,Q)\bigr).
$}
\label{eq:fi}
\end{equation}

\medskip
\noindent\textbf{Definition B.2 (Global Task Mapping).}  
The global mapping \(\hat f\in\mathcal F\) induced by the full sequence \(S^n\) is
\begin{equation}
\resizebox{\columnwidth}{!}{$
\hat f
=\arg\max_{f\in\mathcal F}
\;\mathbb E_{r\sim p_\theta(\cdot\mid S^n)}\bigl[\mathds{1}\{f(\hat I,\hat Q)=r\}\bigr]
\;\Longrightarrow\;
\hat f(\hat I,\hat Q)=\hat R,
$}
\end{equation}
which reduces to
\begin{equation}
\hat f(\hat I,\hat Q)
=\arg\max_{r\in\mathcal R}\;
p_\theta\bigl(r\mid S^n\bigr).
\label{eq:hatf}
\end{equation}

\medskip
\noindent\textbf{Definition B.3 (Mapping Composition).}  
Introduce the composition operator
\(\;C_\theta:\mathcal F^n\times\{\mathrm{Inst}\}\to\mathcal F\)\, such that
\begin{equation}
\hat f
= C_\theta\bigl(f_1,\dots,f_n;\mathrm{Inst}\bigr),
\label{eq:composition}
\end{equation}
capturing how the model integrates local mappings into the global mapping.

\medskip
\noindent\textbf{Definition B.4 (Specific vs.\ Generalized Mapping).}  
\[
\text{Specific‐mapping: }f_1 = f_2 = \cdots = f_n,
\]
\[
\text{Generalized‐mapping: }\exists\,i\neq j,\;f_i\neq f_j.
\]

\section{Vision-language In-context Learning}
\subsection{Demonstration Configuring Details}
\label{demo}
\begin{figure*}
    \centering
    \includegraphics[scale=0.5]{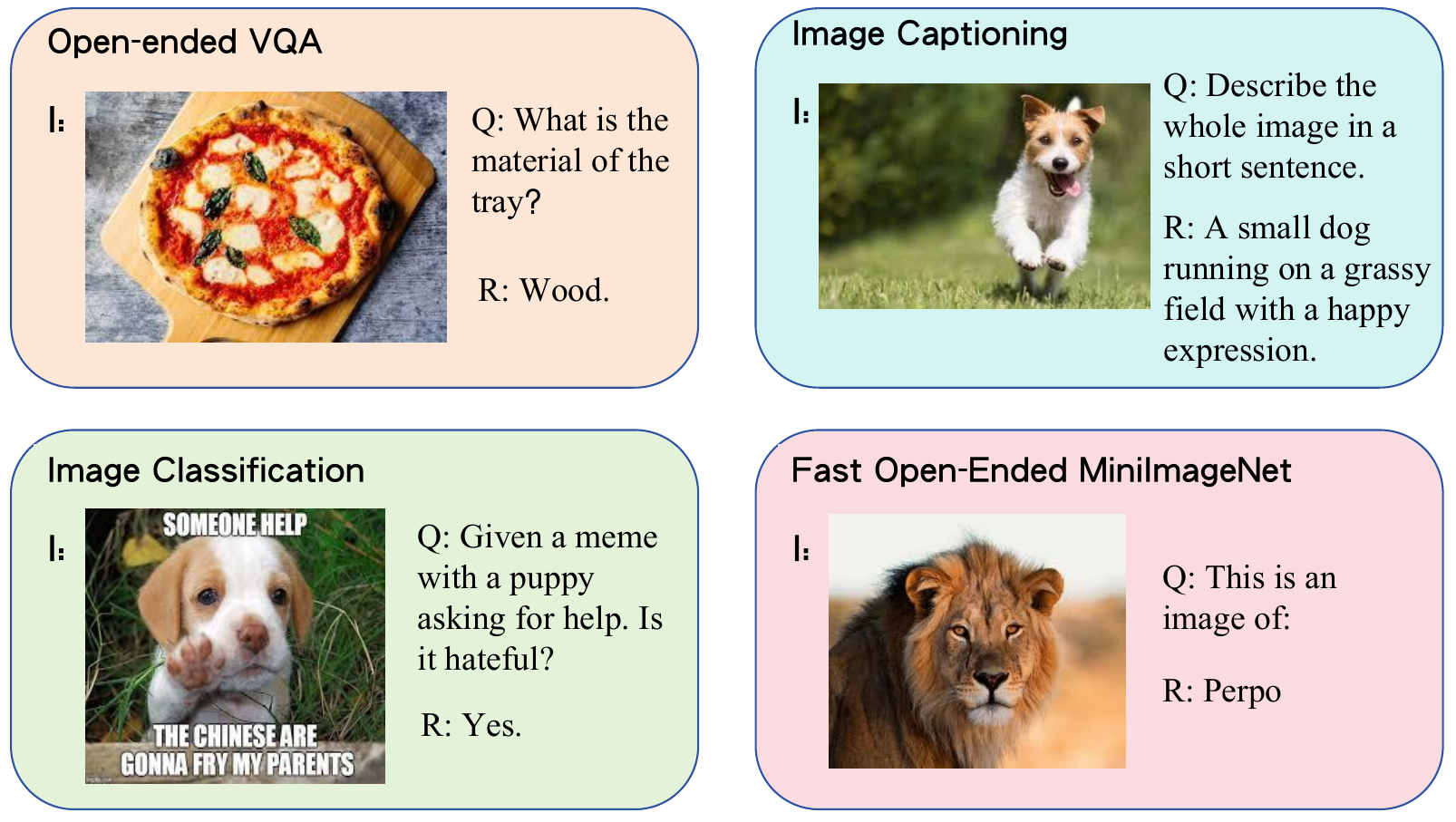}
    \caption{The visualization of (I, Q, R) triplets for Open-ended VQA, image captioning, image classification and Fast Open-ended MiniImageNet.}
    \label{IQR}
\end{figure*}
(a) \textbf{Open-ended VQA}: The query $Q_{i}$ is the single question associated with the image $I_{i}$, while the response $R{i}$ is the answer to the question, provided as a short response. For the query sample, $\hat{Q}$ represents the question related to the image $\hat{I}$, and $\hat{R}$ is the expected output of the model. 

(b) \textbf{Image Captioning}: Both $Q_{i}$ and $\hat{Q}$ are set as short prompts instructing the LVLM to generate a caption for the given image, such as "Describe the whole image in a short sentence.
" The response $R_{i}$ corresponds to the actual caption of the image. 

(c) \textbf{Image Classification}: Both $Q_{i}$ and $\hat{Q}$ provide the textual information paired with the image, followed by a directive requiring the model to classify based on the provided image-text pairs. The response $R_{i}$ is the predefined class label. 

(d) \textbf{Fast Open-ended MiniImageNet}: Both $Q_{i}$ and $\hat{Q}$ are set as short prompts instructing the LVLM to recognize the object in image, such as "This is an image of:" The response $R_{i}$ is the self-defined label. 

(e) \textbf{CLEVR Counting Induction}: Both $Q_{i}$ and $\hat{Q}$ are implicit texts in the form of "attribute: value" pairs. The response $R_{i}$ is the number of objects matching the pairs. 

For all the tasks mentioned above, since the ground-truth answers are not visible to the LVLM during reasoning, all $\hat{R}$ are set to blank. The visualization of (I, Q, R) triplets for the four tasks is shown in Figure \ref{IQR}.
\subsection{In-context Lens}
\label{log}
\begin{figure}[ht]
  \centering
  \includegraphics[width=\columnwidth]{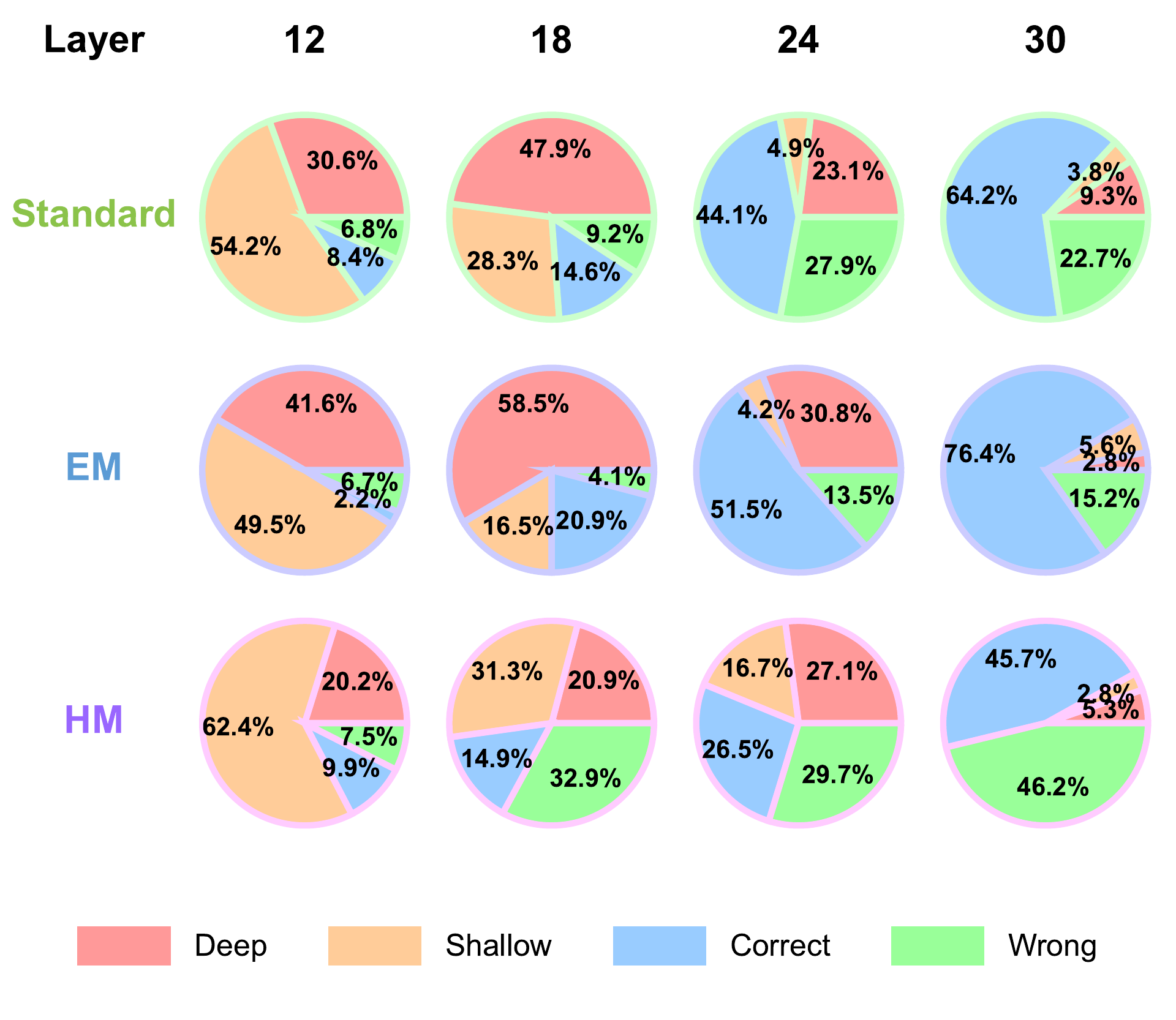}
  \caption{Visualization of the evolving internal representations of an LVLM on the HatefulMemes dataset during ICL, analyzed through the in-context lens.}
  \label{app:pie}
\end{figure}

To visualize how LVLMs’ internal token outputs evolve during ICL on specific-mapping tasks, we introduce the in-context lens, an adaptation of the logit lens \cite{logit}. Like its predecessor, in-context lens projects each layer’s final token embedding back into the text vocabulary. Because local mappings in specific-mapping tasks are largely uniform, we can vary task difficulty to distinguish shallow recognition from deep recognition. In the HatefulMemes example, shallow recognition corresponds to the binary classification decision, while deep recognition requires detecting harmful content. We therefore define four anchor categories by selecting representative keywords: "Shallow" represents superficial task understanding, focusing on general or surface-level concepts. Anchor words include "category," "judge," "label," "identify," and "predict." "Deep" indicates a more profound comprehension of the task, capturing nuanced or context-sensitive meanings. Anchor words include "hateful," "offensive," "biased," "harmful," and "inappropriate." "Correct" corresponds to the correct answer for the query sample. "Wrong" represents the incorrect answer, opposite to "Correct." We then compute, for each layer, the relative probability of the top three relevant decoded tokens falling into each category (summing to 100\%) and visualize the results as pie charts. Figure \ref{app:pie} shows these charts corresponding to the visualizations in Figure \ref{first}(c).
\subsection{Oracle}
\label{self}
Oracle uses the same LVLM $\mathcal{M}$ for both configuring the ICL sequences and performing ICL. This method aims to construct high-quality ICL sequences by iteratively evaluating and selecting demonstrations based on their contribution to the model's predictive performance. Given the ground-truth response $\hat{R} = (\hat{R}^{(1)},..., \hat{R}^{(t)})$ of the query sample, Oracle computes the log-likelihood score $\mathcal{C}_\mathcal{M}(S^n)$ for a sequence $S^n$ with $n$ ICDs, defined as:
\begin{equation}
\mathcal{C}_\mathcal{M}(S^{n})=\sum_{t}^{} log P_{\mathcal{M}}(\hat{R}^{(t)} \mid S^{n},\hat{R}^{(1:t-1)}),
\end{equation}
where $\mathcal{M}$ denotes the LVLM. This score measures how effectively the model predicts the ground-truth response $\hat{R}$ given the current ICL sequence $S^n$.

The configuration process begins with an empty sequence $S^0$ and iteratively selects demonstrations. At each step $n$, a demonstration $x_n$ is chosen from the library $D$ to maximize the incremental gain in the log-likelihood score:
\begin{equation}
x_{n}= \underset {x\in D}{argmax}[\mathcal{C
}_\mathcal{M}(S^{n-1}+x)-\mathcal{C
}_\mathcal{M}(S^{n-1})].
\end{equation}
This greedy optimization process ensures that each selected demonstration contributes optimally to the sequence. Unlike simple similarity-based methods, Oracle evaluates the overall impact of each candidate demonstration on the sequence's quality.
\subsection{Ablation Settings}
\label{setting}
To systematically evaluate the impact of task mapping in multimodal in-context learning (ICL), we design controlled ablation settings that selectively perturb key factors such as \textbf{label reliability} and \textbf{visual modality}. Below, we provide detailed descriptions of each setting's implementation.
\begin{enumerate}[label=\arabic*., leftmargin=1em]
  \item \textbf{Label Reliability}
    \begin{itemize}[noitemsep, leftmargin=0.5em]
      \item \textit{Wrong Labels (WL)}: To evaluate the reliance on explicit label correctness, we invert 75\% \(R_i\) labels (yes\(\leftrightarrow\)no) in the ICL sequence. This setting disrupts direct label-based learning while maintaining the overall task structure, allowing us to examine whether LVLMs primarily depend on task mapping rather than correct labels.
    \end{itemize}

  \item \textbf{Visual modality}
    \begin{itemize}[noitemsep, leftmargin=0.5em]
      \item \textit{Blur Images (BI)}: To investigate the role of visual information clarity, we apply Gaussian blur to the images \(I_i\) in the ICL sequence. This degrades fine-grained details while preserving overall structure, allowing us to examine the impact of visual degradation on task mapping.
      \item \textit{BI on Query Image (BI($\hat{I}$))}: Instead of applying blur to the entire ICL sequence, (BI)$\hat{I}$ applies Gaussian blur only to the query image $\hat{I}$. This setting helps isolate the effect of degraded query information on task mapping performance.
    \end{itemize}
    
  \item \textbf{Query Enhancement}
    \begin{itemize}[noitemsep, leftmargin=0.5em]
      \item \textit{Easier Mapping on Query (EM($\hat{Q}$))}: This setting enhances the query text $\hat{Q}$ by incorporating explicit task guidance to facilitate task mapping. Instead of modifying the ICL sequence, EM($\hat{Q}$) provides additional textual hints that reinforce task semantics, allowing us to measure whether improved query understanding compensates for suboptimal ICD configurations.
    \end{itemize}
\end{enumerate}

\subsection{Task Mapping Cohesion Metrics}
\label{sensitive}

\subsubsection{Disruption Gap (\texorpdfstring{$\Delta$}{Delta})}
To measure the impact of individual ICDs on sequence-level performance and assess task mapping cohesion, we define the Disruption Gap ($\Delta$) as the magnitude of performance change caused by replacing a single ICD in the sequence.

For each ICD $x_i = (I_i, Q_i, R_i)$ in the sequence $S^n$, a replacement ICD $x_j = (I_j, Q_j, R_j)$ is selected from the same dataset based on the highest joint similarity of their image and query embeddings (IQ2IQ). The modified sequence $S_{\text{replaced}, i}$ is then constructed by replacing $x_i$ with $x_j$.

The Disruption Gap for the $i$-th ICD is defined as the absolute difference in performance before and after the replacement:

\begin{equation}
\Delta_i = \big|\mathcal{L}(S) - \mathcal{L}(\mathcal{S}_{\text{replaced}, i})\big|,
\end{equation}
where $\mathcal{L}(\cdot)$ represents the performance metric of the sequence (e.g., accuracy).

For a sequence $\mathcal{S}$ with $N$ ICDs, the overall Disruption Gap is computed as the average $\Delta_i$ across all $N$ ICDs:
\begin{equation}
\Delta = \frac{1}{N} \sum_{i=1}^N \Delta_i.
\end{equation}

To ensure the robustness of $\Delta$ and to account for potential variability in replacement effects, we conduct repeated experiments. This metric quantifies the sequence's cohesion by assessing the sensitivity of the overall performance to individual replacements. A higher $\Delta$ indicates that the sequence has stronger cohesion, as replacing an ICD results in larger performance changes.
\subsubsection{Order Sensitivity (\texorpdfstring{$\sigma$}{sigma})}
For an ICL sequence $S^{n}$, we generate $K$ independent random permutation of it:
\begin{equation}
{S^n_{\text{permute},1}, S^n_{\text{permute},2}, \ldots, S^n_{\text{permute},K}}, \quad K = 10.
\end{equation}
Then we compute the accuracy for each permuted sequence $k = 1, 2, \ldots, K$:

\begin{align}
\text{Acc}\bigl(S^n_{\text{permute},k}\bigr)
  &= \frac{\text{Correct Predictions}}{\text{Total Predictions}}.
\end{align}

Then calculate the mean accuracy across all permutations:
\begin{equation}
\mu = \frac{1}{K} \sum_{k=1}^K \text{Acc}(S^n_{\text{permute},k)}.
\end{equation}
Finally, compute the standard deviation of accuracies as $\sigma$:
\begin{equation}
\sigma = \sqrt{\frac{1}{K} \sum_{k=1}^K \left( \text{Acc}(S^n_{\text{permute},k) }- \mu \right)^2}.
\end{equation}

\subsubsection{Metric Analysis}
$\Delta$ and $\sigma$ together constitute a rigorous framework for quantifying the cohesion of task mappings in in-context learning. The Disruption Gap is defined as the mean absolute degradation in task performance when each ICD is replaced by its nearest neighbor in the learned representation space; this metric directly captures the indispensability of individual local mappings for preserving the semantics of the overall task. A larger value of $\Delta$ implies that each ICD contributes uniquely and cannot be substituted without harming global inference. Order Sensitivity is computed as the standard deviation of task accuracy across multiple random permutations of ICD order; this metric assesses the invariance of the global mapping to the structural arrangement of examples. A smaller value of $\sigma$ indicates that the inferred mapping remains stable regardless of ICL sequence, reflecting intrinsic consistency among local mappings. By combining $\Delta$ and $\sigma$, one obtains a complementary view in which $\Delta$ measures discriminative necessity and $\sigma$ measures structural resilience, thus ensuring that a truly cohesive task mapping exhibits both strong local-to-global alignment and robustness to variations in ICD composition.  
\subsection{Case Study}
\label{case}
\begin{figure*}
    \centering
    \includegraphics[width=\textwidth]{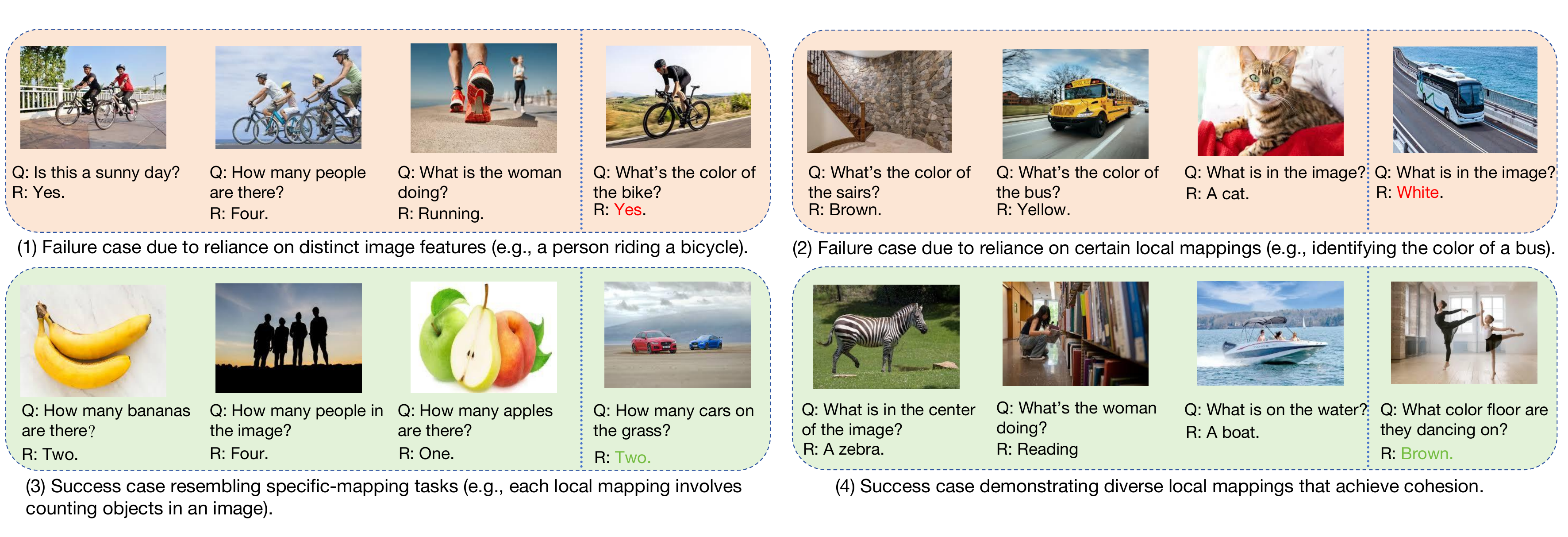}
    \caption{Four types of ICL sequences in the generalized-mapping tasks. }
    \label{ca}
\end{figure*}

In Figure \ref{ca}, we present four examples representing the four typical types of ICL sequences in generalized-mapping tasks.
\section{Method}
\subsection{CLIP Encoders}
CLIP employs two distinct encoders: one for images and another for text. The image encoder transforms high-dimensional visual data into a compact, low-dimensional embedding space, using architectures such as a ViT. Meanwhile, the text encoder, built upon a Transformer architecture, generates rich textual representations from natural language inputs.

CLIP is trained to align the embedding spaces of images and text through a contrastive learning objective. Specifically, the model optimizes a contrastive loss that increases the cosine similarity for matched image-text pairs, while reducing it for unmatched pairs within each training batch. To ensure the learning of diverse and transferable visual concepts, the CLIP team curated an extensive dataset comprising 400 million image-text pairs, allowing the model to generalize effectively across various downstream tasks. 

In our experiments, we employ the same model, CLIP-ViT-L/14, using its image and text encoders to generate the image and text embeddings for each demonstration, ensuring consistency in cross-modal representations. The model employs a ViT-L/14 Transformer architecture as the image encoder and a masked self-attention Transformer as the text encoder. We experimented with several strategies for training the CLIP encoder and found that training only the last three layers of the encoder offers the best cost-effectiveness.

\subsection{Instruction}
\label{instapp}
The $Inst$ generated by GPT-4o in the main experiment is "You will be provided with a series of image-text pairs as examples and a question. Your task involves two phases: first, analyze the provided image-text pairs to grasp their context and try to deeply think about what the target task is; second, use this understanding, along with a new image and your knowledge, to accurately answer the given question." This content demonstrates great orderliness and can act as a good general semantic guide for ICDs and the query sample. This style is named chain-of-thought (CoT) \cite{stare}.

To incorporate the semantic information of $Inst$ and strengthen task representation during the ICL sequence configuration process, we use GPT-4o to generate simplified versions of these $Inst$ and integrate their embeddings into the task guider, which are indicated by $Inst'$. The prompt we use is as follows: \textit{"This is an instruction to enable LVLMs to understand and perform a multimodal in-context learning task. Please simplify it by shortening the sentence while preserving its function, core meaning, and structure. The final version should be in its simplest form, where removing any word would change its core meaning."} This simplification process allows us to investigate how the semantic information density in the instruction impacts TACO's sequence configuration ability and the performance of LVLMs in ICL. The results show that simplifying the instruction in a prompt before embedding it in the task guider significantly improves the quality of sequence generation. It also helps to avoid issues caused by too long instructions. 

As shown in Table \ref{instde}, we use GPT-4o to rewrite $Inst$, placing it at the middle and the end of a prompt, altering its semantic structure accordingly while keeping its CoT nature. The table also presents two other tested styles of instructions placed at the beginning of the prompt: Parallel Pattern Integration (PPI) and System-Directive (SD). PPI emphasizes simultaneous processing of pattern recognition and knowledge integration, focusing on dynamic pattern repository construction rather than sequential reasoning. SD structures input as a formal system protocol with defined parameters and execution flows, prioritizing systematic processing over step-by-step analysis. These two forms have also been proven to be effective in previous ICL work. We use them to study the robustness of TACO and various LVLMs to different instruction formats.
\begin{table*}

\centering
\resizebox{\textwidth}{!}{\begin{tabular}{m{4.5cm}<{\centering} | m{10cm} }
\toprule
\textbf{$Inst$} & \textbf{Details} \\
\midrule
\textit{Beginning1} (CoT)& You will be provided with a series of image-text pairs as examples and a question. Your task involves two phases: first, analyze the provided image-text pairs to grasp their context and try to deeply think about what the target task is; second, use this understanding, along with a new image and your knowledge, to accurately generate the response to the given query. \\

\midrule
\textit{Beginning2} (PPI)& Construct a dynamic pattern repository from image-text samples, then leverage this framework alongside your knowledge base for concurrent visual analysis and query resolution. The key is parallel processing - your pattern matching and knowledge integration should happen simultaneously rather than sequentially. \\

\midrule
\textit{Beginning3} (SD)& SYSTEM DIRECTIVE
Input Stream: Example Pairs → New Image + Query
Process: Pattern Extract → Knowledge Merge → Visual Analysis → Response Critical: All exemplar patterns must inform final analysis Priority: Context preservation essential\\

\midrule
\textit{Middle} (CoT)& Now you have seen several examples of image-text pairs. Next, you will be given a question. Your task involves two phases: first, revisit the above image-text pairs and try to deeply think about what the target task is; second, use this understanding, along with a new image and your knowledge, to accurately generate the response to the given question. \\

\midrule
\textit{End} (CoT)& Now you have seen several examples of image-text pairs and a question accompanied by a new image. Your task involves two phases: first, revisit the provided examples and try to deeply think about what the target task is; second, use this understanding, the new image, and your knowledge to accurately generate the response of the given question. \\

\midrule
\textit{Beginning1 (Abbreviated)} & Analyze the following image-text pairs, understand the task, and use this to generate the response with a new image. \\
\midrule
\textit{Middle (Abbreviated)} & After reviewing the above image-text pairs, analyze the task and use this understanding to generate the response with a new image. \\
\midrule
\textit{End (Abbreviated)} & After reviewing the above image-text pairs and a query with a new image, analyze the task and use this understanding to generate the response. \\

\bottomrule

\end{tabular}}
\caption{\label{instde} Formats of different instruction types and their corresponding details used in the prompt structure for all VL tasks. (Abbreviated) means that the instruction is a simplified version produced by GPT-4o. }

\end{table*}

\subsection{Prompt Details}
\label{prompt}
The prompts constructed based on $S^{n}$ all follow the format:
$$(Inst; ICD_{1}, ..., ICD_{n}; Query Sample).$$ 
Each ICD's query begins with "Query:" and its response starts with "Response:". The query sample concludes with "Response:", prompting the LVLM to generate a response. Depending on the input format required by different LVLMs, we may also include special tags at the beginning and end of the prompt. 

Each model, including OpenFlamingov2, Idefics2, InternVL2.5, and Qwen2.5VL, employs a structured approach to engage with image-text pairs. The two-phase task requires LVLMs to first absorb information from a series of prompts before utilizing that context to answer subsequent questions related to new images. This method allows for enhanced understanding and reasoning based on prior knowledge and context, which is essential for accurate predictions in VL tasks.

\section{Experiment}
\subsection{Datasets and Models}
\label{dL}
\subsubsection{Dataset}

\begin{figure*}
    \centering
    \includegraphics[scale=0.55]{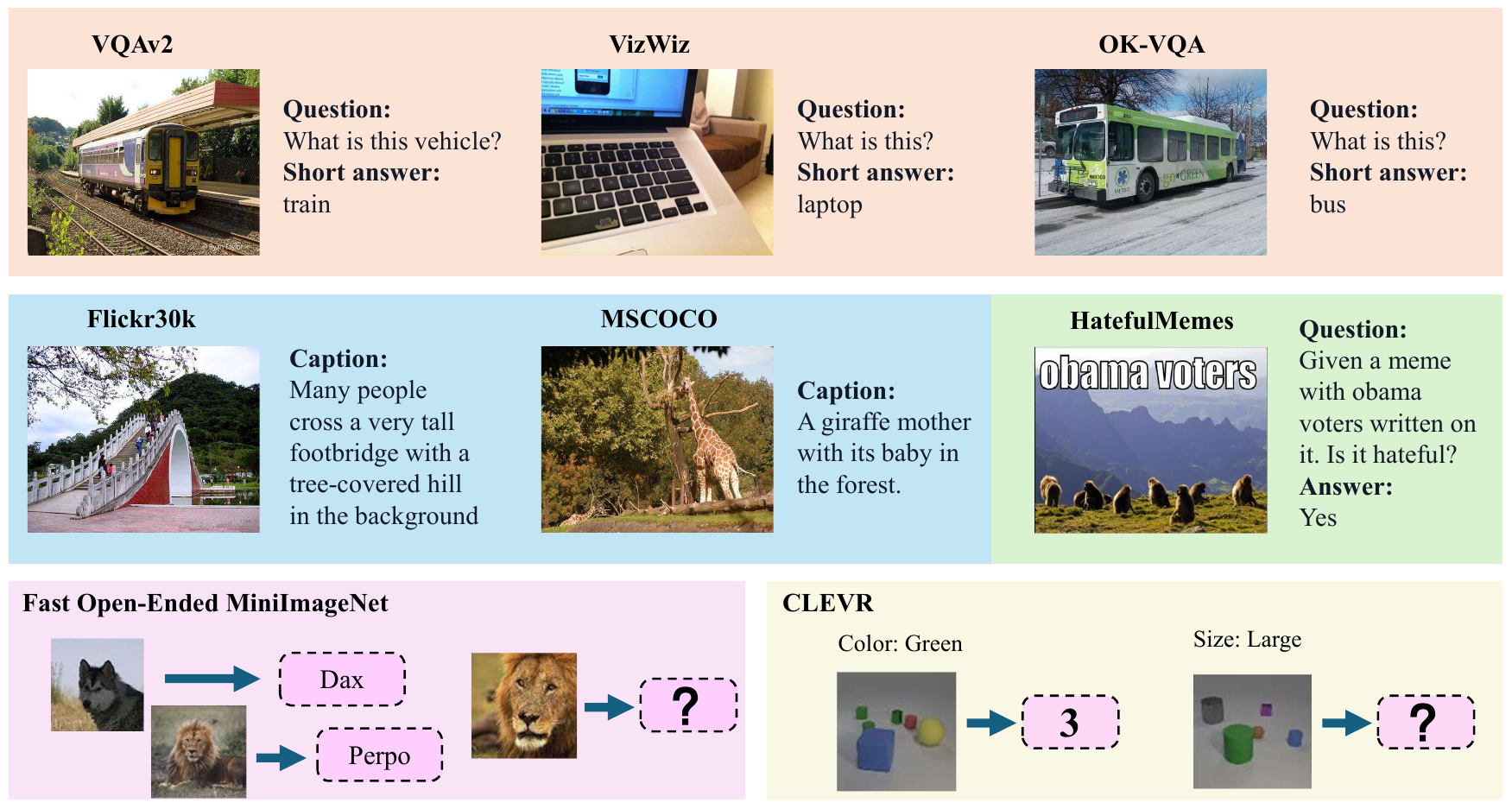}
    \caption{Illustrative examples from various vision-and-language datasets categorized by task type. Visual Question Answering (VQA) tasks are shown in red (VQAv2: train, VizWiz: laptop, OK-VQA: bus). Captioning tasks are represented in blue (Flickr30k: footbridge, MSCOCO: giraffes), while classification tasks are highlighted in green (HatefulMemes: meme identified as hateful). The bottom section demonstrates reasoning tasks with synthetic datasets: Fast Open-Ended MiniImageNet and CLEVR, focusing on conceptual understanding (e.g., assigning labels like "Dax" or identifying object properties like color and size).}
    \label{datasets}
\end{figure*}
In our study, we explore various VL tasks that use diverse datasets to evaluate model performance. As illustrated in Figure \ref{datasets}, we use VQA datasets such as VQAv2, VizWiz, and OK-VQA, which test the models' abilities in question-answer scenarios. Additionally, we incorporate image captioning datasets such as Flickr30k and MSCOCO to assess descriptive accuracy, along with the HatefulMemes dataset for classification tasks focused on hate speech detection. This comprehensive approach allows us to thoroughly evaluate the models across different tasks. The size distribution of the training, validation and test sets in these VL datasets is shown in Table \ref{size}.

\begin{table}
\centering
\resizebox{\columnwidth}{!}{\begin{tabular}{c c c c c}
\toprule
\textbf{Datasets} & \textbf{Training} & \textbf{Validation} & \textbf{Test} & \textbf{$\hat{D}$ Size}\\
\midrule
VQAv2 & 443,757 & 214,354 & 447,793& 8000\\
VizWiz & 20,523 & 4,319 & 8,000&2000\\
OK-VQA & 9,055 & 5,000 & /&800\\
Flickr30k & 29,783 & 1,000 & 1,000&2500\\
MSCOCO & 82,783 & 40,504 & 40,775&3000\\
HatefulMemes & 8,500 & 500 & 2,000&800\\
\textbf{Hybrid} & 30000 & 9000 & /& 3000\\
\textbf{Fast} & 5,000 & / & 200& 500\\
\textbf{CLEVR} & 800 & / & 200& 80\\
\bottomrule
\end{tabular}}
\caption{\label{size} Overview of the size distribution across the datasets used.}
\end{table}

For the Open-ended VQA task, we utilize the following datasets: VQAv2, which contains images from the MSCOCO dataset and focuses on traditional question-answering pairs, testing the model's ability to understand both the image and the question. VizWiz presents a more challenging setting with lower-quality images and questions, along with a lot of unanswerable questions, pushing models to handle uncertainty and ambiguity. OK-VQA is distinct in that it requires the model to leverage external knowledge beyond the image content itself to generate correct answers, making it a benchmark for evaluating models’ capacity to integrate outside information.

For the Image Captioning task, we use the Flickr30k and MSCOCO datasets. The Flickr30k dataset consists of images depicting everyday activities, with accompanying captions that provide concise descriptions of these scenes. The MSCOCO dataset is a widely-used benchmark featuring a diverse range of images with detailed and richly descriptive captions, ideal for evaluating image captioning models.

For the Image Classification task, we use the HatefulMemes dataset, which is an innovative dataset designed to reflect real-world challenges found in internet memes. It combines both visual and textual elements, requiring the model to jointly interpret the image and the overlaid text to detect instances of hate speech.

VL-ICL Bench covers a number of tasks, which include diverse multimodal ICL capabilities spanning concept binding, reasoning or fine-grained perception. Few-shot ICL is performed by sampling the ICDs from the training split and the query examples from the test split. We choose two image-to-text generation tasks from it, which reflects different key points of ICL. Fast Open MiniImageNet task assigns novel synthetic names (e.g., dax or perpo) to object categories, and LVLMs must learn these associations to name test images based on a few examples instead of their parametric knowledge, emphasizing the importance of rapid learning from ICDs. CLEVR Count Induction asks LVLMs to solve tasks like \textit{"How many red objects are there in the scene?"} from examples rather than explicit prompts. The ICDs' images are accompanied by obscure queries formed as attribute-value pairs that identify a specific object type based on four attributes: size, shape, color, or material. Models must perform challenging reasoning to discern the task pattern and generate the correct count of objects that match the query attribute.

The datasets in our experiments are evaluated using task-specific metrics, as summarized in Table \ref{metrics}. For the VQA tasks, \textbf{Hybrid} dataset and tasks in VL-ICL Bench, we use accuracy as the metric to assess the models' ability to provide correct answers.

\begin{table*}[t]
\centering
\resizebox{\textwidth}{!}{\begin{tabular}{c | c c c c c c c c c}
\toprule
\textbf{Datasets} & \textbf{VQAv2} & \textbf{VizWiz} & \textbf{OK-VQA} & \textbf{Flickr30k} & \textbf{MSCOCO} & \textbf{HatefulMemes} & \textbf{Hybrid} & \textbf{Fast} & \textbf{CLEVR}\\
\midrule
metrics & Accuracy & Accuracy & Accuracy & CIDEr & CIDEr & ROC-AUC & Accuracy & Accuracy & Accuracy\\
\bottomrule
\end{tabular}}
\caption{\label{metrics} Evaluation metrics used for each benchmark. Accuracy is used for VQA datasets (VQAv2, VizWiz, OK-VQA), self-built \textbf{Hybrid} dataset, and two tasks in VL-ICL Bench. CIDEr \cite{cidervedantam2015cider} is used for image captioning datasets (Flickr30k, MSCOCO). ROC-AUC is used for the HatefulMemes classification task.}

\end{table*}
For the image captioning tasks, we use the CIDEr score, which measures the similarity between generated captions and human annotations. Finally, for the HatefulMemes classification task, we evaluate performance using the ROC-AUC metric, which reflects the model's ability to distinguish between hateful and non-hateful content.

\subsubsection{LVLMs}
In recent advances of LVLMs, efficient processing of multimodal inputs, especially images, has become a critical focus \cite{pano,mmsurvey,cogvla,breaking}. Models like OpenFlamingov2, Idefics2, InternVL2.5, Qwen2.5VL, and GPT-4V implement unique strategies to manage and process visual data alongside textual input \cite{relayer,fairreason,forgetting,mc}.

OpenFlamingov2 handles visual input by dividing images into patches and encoding them with a Vision Transformer. Each image patch generates a number of visual tokens, which are then processed alongside text inputs for multimodal tasks. To manage multi-image inputs, the model inserts special tokens <image> and <|endofchunk|> at the beginning and end of the visual token sequences. For example, an image divided into 4 patches produces 4 x 256 visual tokens, with the additional special tokens marking the boundaries before the tokens are processed by the large language model. 

Idefics2 processes visual input by applying an adaptive patch division strategy adapted to image resolution and content complexity. Depending on these factors, each image is segmented into 1 to 6 patches, striking a balance between preserving spatial information and maintaining efficiency. These patches are encoded through a Vision Transformer, followed by a spatial attention mechanism and a compact MLP, resulting in 128 visual tokens per patch. The positions of images in the input sequence are marked with <|image\_pad|> for alignment, while <end\_of\_utterance> tokens separate query and answer components in in-context demonstrations. An image split into five patches yields 5 x 128 + 2 tokens before being integrated with the LLM. 

InternVL2.5 dynamically divides each input image into tiles by selecting the closest aspect ratio i/j from a predefined set and resizing the image to S×i by S×j (with S = 448) before splitting it into i×j non-overlapping 448×448 px patches. Each patch is then fed through the InternViT encoder (InternViT-300M or InternViT-6B) to produce 1 024 patch embeddings, which are spatially downsampled via a pixel-unshuffle operation to yield exactly 256 visual tokens per patch. Special <img> and </img> tokens are inserted at the start and end of the full token sequence, so an image split into 3 patches produces 3 × 256 + 2 tokens before being passed to the LLM.

Qwen2.5VL reduces the number of visual tokens per image via an MLP-based merger that concatenates and compresses spatially adjacent patch features. A native-resolution ViT first splits an image (e.g. 224 × 224 with patch size 14) into a grid of patch embeddings. Rather than feeding all raw patches into the LLM, Qwen2.5VL groups each 2 × 2 block of adjacent patch features (four tokens), concatenates them, and projects the result through a two-layer MLP into a single fused token aligned with the LLM’s embedding dimension. This achieves a 4× reduction in sequence length, dynamically compressing image feature sequences of varying lengths.

GPT-4V (Vision) extends GPT-4's capabilities to handle VL tasks by enabling the model to process and reason about visual input alongside text. The model can perform various tasks including image understanding, object recognition, text extraction, and visual question-answering through natural language interaction. In terms of its few-shot learning ability, GPT-4V demonstrates the capacity to adapt to new visual tasks given a small number of examples through natural language instructions, showing potential in areas such as image classification and visual reasoning, though performance may vary across different task domains and complexity levels.

\subsection{Training Data Construction Details}
\label{training}
We construct sequence data for model training using existing high-quality datasets, each corresponding to a VL task. The samples are uniformly formatted as $ (I, Q, R) $ triplets based on their respective task types. Each dataset generates a sequence set $ D_S $ for training, where each sequence consists of a query sample and $ N $ ICDs. The value of $ N $ is configurable, determining the number of shots during training. To ensure optimal training performance, we employ the same LVLM used in inference as a scorer to supervise the construction of $ D_S $, making the method inherently model-specific. For each dataset, we construct $ D_S $ exclusively from its training set through the following three-step process: (1). We apply $ k $-means clustering based on image features to partition the dataset into $ k $ clusters. From each cluster, we select the $ m $ samples closest to the centroid, yielding a total of $ K = m \times k $ samples. These form the query sample set $ \hat{D} $ after removing their ground-truth responses, which are stored separately in $ D_{\hat{R}} $. The remaining dataset serves as the demonstration library $ DL $. (2). For each query sample $ \hat{x}_i \in \hat{D} $, we randomly sample a candidate set $ D_i $ of $ 64n $ demonstrations from $ DL $. The objective is to retrieve $ N $ demonstrations from $ D_i $ that optimally configure the sequence for $ \hat{x}_i = (\hat{I}_i, \hat{Q}_i) $ with its ground-truth response $ \hat{R}_i = (\hat{R}_i^{(1)},..., \hat{R}i^{(t)}) $. We use the log-likelihood score computed by the LVLM $ \mathcal{M} $ as the selection criterion $ \mathcal{C}\mathcal{M} $, evaluating the model's predictive ability given a sequence with $n$ ICDs:
\begin{equation}
\mathcal{C}_\mathcal{M}(S^{n}_{i})=\sum_{t}^{} log P_{\mathcal{M}}(\hat{R}_{i}^{(t)} \mid S^{n}_{i},\hat{R}_{i}^{(1:t-1)}),
\end{equation}

To determine the optimal $ n $-th demonstration $ x_n $ for a sequence $ S_i^{n-1} $ with $ n-1 $ ICDs, we select the candidate that maximizes the incremental gain in $ \mathcal{C}_\mathcal{M} $:

\begin{equation}
x_{n}= \underset {x\in D_{i}}{argmax}[{\mathcal{C}_\mathcal{M}}(S^{n-1}_{i}+x)-{\mathcal{C}_\mathcal{M}}(S^{n-1}_{i})].
\end{equation}

(3). We employ beam search with a beam size of $ 2N $, ensuring that for each $ \hat{x} $, the top $ 2N $ optimal sequences are included in $ D_S $. As a result, the final sequence set $ D_S $ consists of $ 2N \times k $ $ N $-shot sequences, providing refined training data for the model.
\subsection{Baselines}
\label{baseline}
Various baseline methods are used to evaluate the model's performance, ranging from random sampling to different SOTA retrieval strategies. The following is a description of the baselines used in our experiments.

1. \textbf{Random Sampling (\textbf{RS})}: In this approach, a uniform distribution is followed to randomly sample $n$ demonstrations from the library. These demonstrations are then directly inserted into the prompt to guide the model in answering the query.

2. \textbf{Image2Image (I2I)}: During the retrieval process, only the image embeddings $I_{i}$ from each demonstration $(I_{i}, Q_{i}, R_{i}$ are used. These embeddings are compared to the query image embedding $\hat{I}$ and the retrieval is based on the similarity between the images.

3. \textbf{ImageQuery2ImageQuery (IQ2IQ)}: During the retrieval process, both the image embeddings $I_{i}$ and the query embeddings $Q_{i}$ of each demonstration $(I_{i}, Q_{i}, R_{i}$ are used. These embeddings are compared to the embedding of the concatenated query sample $(\hat{I},\hat{Q})$, and the retrieval is based on the joint similarity between the images and the queries.

4. \textbf{ImageQuery\&Pseudo Response (IQPR)}: This baseline begins by using RS to generate a pseudo response $\hat{R}^{P}$ for the query sample. The pseudo response is concatenated with $\hat{I}$ and $\hat{Q}$ to create the query sample’s complete embedding. We then retrieve $4n$ candidate samples from the dataset based on the similarity to this full embedding, and finally select the top $n$ ICDs from these candidates using their Q–R similarity.

5. \textbf{DEmO}: DEmO is a two‐stage, data‐free framework for configuring an optimal in‐context sequence using influence, a concept that has become increasingly popular in ICD selection. In the first stage, it draws \(N\) random permutations \(\{\pi_i\}_{i=1}^N\) of the candidate support set and measures each permutation’s label‐fairness by computing its content‐free entropy:
\begin{equation}
  E(\pi)\;=\; -\sum_{l} P\bigl(y=l\mid C_{\pi}\bigr)\,\log P\bigl(y=l\mid C_{\pi}\bigr),
\end{equation}
where \(C_{\pi}\) is the prompt constructed by \(\pi\) with a content‐free token in place of the query. The top-\(K\) permutations with the highest \(E(\pi)\) are retained as the candidate set \(\Pi\).

In the second stage, for each candidate \(\pi\in\Pi\) and test input \(x_t\), DEmO computes the influence score
\begin{equation}
\begin{split}
  I(x_t;\pi) &= P\bigl(y^*\mid x_t, C_{\pi}\bigr) 
             -\,P\bigl(y^*\mid C_{\pi}\bigr),\\
  y^*        &= \arg\max_y P\bigl(y\mid x_t, C_{\pi}\bigr),
\end{split}
\end{equation}
which quantifies how much adding \(x_t\) shifts the model’s confidence in its most likely label. 

Finally, permutation \(\pi^*=\arg\max_{\pi\in\Pi}I(x_t;\pi)\) is chosen for the actual prediction. This targeted re‐ranking ensures that each test sample uses the example order most “influential” to its correct classification, without relying on any additional labeled data.

6. \textbf{Lever-LM}: Lever-LM is designed to capture statistical patterns between ICDs for an effective ICL sequence configuration. Observing that configuring an ICL sequence resembles composing a sentence, Lever-LM leverages a temporal learning approach to identify these patterns. A special dataset of effective ICL sequences is constructed to train Lever-LM. Once trained, its performance is validated by comparing it with similarity-based retrieval methods, demonstrating its ability to capture inter-ICD patterns and enhance ICL sequence configuration for LVLMs.

\begin{table}
\resizebox{\columnwidth}{!}{\begin{tabular}{c c c c c c}
\toprule
\textbf{Datasets} & \textbf{Training} & \textbf{Validation} & \textbf{Test} & \textbf{$\hat{D}$ Size} & metrics\\
\midrule
Rule Learning & 1600 & - & 150& & exact match scores\\
Fast Counting & 800 & - & 40& & Accuracy\\
\bottomrule
\end{tabular}}
\caption{\label{size1} Overview of Rule Learning and Fast Counting tasks.}
\end{table}

\subsection{Results and Analysis}
\label{mainapp}
\begin{table*}
\centering
\resizebox{\textwidth}{!}{\begin{tabular}{c | c c c c | c c | c | c | c | c}
\toprule
\multirow{2}{*}{\textbf{Model}} & \multirow{2}{*}{\textbf{Method}}&\multicolumn{3}{c|}{\textbf{VQA}} & \multicolumn{2}{c|}{\textbf{Captioning}} & \multicolumn{1}{c|}{\textbf{Classification}} & \multirow{2}*{\textbf{Hybrid}} & \multirow{2}*{\textbf{Fast}} & \multirow{2}*{\textbf{CLEVR}}\\
\cmidrule(lr){3-8}
& & VQAv2 & VizWiz & OK-VQA &  Flickr30K & MSCOCO  & HatefulMemes &~ & ~& ~\\
\midrule
\multirow{7}*{OpenFlamingov2} & RS & 50.84 & 27.71 & 37.90 & 76.74 & 92.98 & 64.75 & 13.48 & 57.69 & 21.60 \\

~ & I2I & 49.52& 26.82 & 37.79 & 79.84 & 94.31 & 69.53 & 12.79 & 59.07 & 19.39 \\

~ & IQ2IQ & 52.29 & 31.78 & 42.93 & 79.91 & 94.40 & 68.72 & 24.93 & 58.96 & 20.03\\

~ & IQPR & 53.38 & 30.12 & 41.70 & 80.02 & 96.37 &  69.16 & 28.71 & 57.32 & 21.84 \\

~ & DEmO & 51.34 & 32.09 & 42.88 & 81.25 & 95.70 & 65.87 & 25.97 & 58.49 & 20.69 \\

~ & Lever-LM & 55.89 & 33.34 & 43.65 & 83.17 & 98.74 & 72.70 & 32.04 & 59.41 & 22.67\\

~ & Ours & \textbf{61.12} & \textbf{39.76} & \textbf{47.28} & \textbf{84.23} & \textbf{99.10} & \textbf{75.09} & \textbf{35.17}& \textbf{60.25} & \textbf{24.80} \\
\midrule
\multirow{7}*{Idefics2} & RS & 54.97 & 32.92 & 40.01 & 82.43 & 99.61 & 69.31 & 15.65 & 54.72 & 35.14 \\

~ & I2I & 53.77 & 31.67 & 41.37 & 85.76 & 101.34 & 69.64 & 10.49 & 55.20 & 32.37 \\

~ & IQ2IQ & 55.41 & 34.31 & 43.13 & 85.63 & 101.45 & 70.78 & 30.36 & 55.14 & 32.75\\

~ & IQPR & 55.32 & 33.74 & 42.76 & 87.65 & 103.57 & 62.18 & 24.03 & 55.18 & 36.29\\

~ & DEmO & 54.01 & 35.12 & 42.87 & 87.83 & 104.31 &  68.52 & 23.76 & 54.09 & 37.13 \\

~ & Lever-LM & 56.78 & 34.10 & 43.27 & 88.01 & 105.62 & 71.33 & 30.14 & 55.83 & 38.97 \\

~ & Ours & \textbf{59.41} & \textbf{38.32} & \textbf{48.35} & \textbf{90.41} & \textbf{107.04} & \textbf{73.68} & \textbf{33.25}&\textbf{57.21} & \textbf{40.21} \\
\midrule
\multirow{7}*{InternVL2.5} & RS & 66.73 & 56.54 & 59.85 & 102.37 & 119.26 & 73.82 & 19.03 & 75.79 & 58.82 \\

~ & I2I & 64.71 & 56.03 & 59.51 & 105.31  & 121.10& 77.05 & 16.03 & 77.03 & 57.79 \\

~ & IQ2IQ & 68.92 & 57.86 & 64.19 & 105.33 & 124.36 & 79.95 & 40.82 & 79.35 & 56.48 \\

~ & IQPR & 70.01 & 58.19 & 67.58 & 106.52 & 125.73 & 81.20 & 42.39 & 79.41 & 60.42\\

~ & DEmO & 69.58 & 56.37 & 68.64 & 105.85 & 123.94 & 82.16 & 41.79 & 78.62 & 56.37 \\

~ & Lever-LM & 72.61 & 59.45 & 70.28 & 106.32 & \textbf{127.51} & 82.04 & 45.77 & 80.72 & 62.08 \\

~ & Ours & \textbf{74.82} & \textbf{62.73} & \textbf{73.05}& \textbf{109.16} & 127.43 & \textbf{84.72} & \textbf{47.39} &\textbf{81.61}  & \textbf{64.15} \\
\midrule
\multirow{7}*{Qwen2.5VL} & RS & 68.59 & 54.37 & 62.38 & 105.26 & 126.32 & 80.41 & 23.58 & 73.26 & 56.48\\

~ & I2I & 66.98 & 53.81 & 62.75 & 105.78 & 126.43 & 78.62 & 15.79 & 79.84 & 54.83\\

~ & IQ2IQ & 68.85 & 55.87 & 65.37 & 106.07 & 127.95 & 79.89 & 43.28& 79.57 & 57.06\\

~ & IQPR & 70.28 & 57.92 & 66.28 & 106.57 & 128.42 & 81.96 & 47.38 & 78.82 & 57.37\\

~ & DEmO & 69.47 & 58.06 & 66.75 & 105.92 & 129.01 & 79.53 & 46.73 & 77.61 & 54.28 \\

~ & Lever-LM & 70.06 & 59.16 & 68.72 & 107.35 & 132.48 & 83.42 & 54.47 & \textbf{80.53} & 60.47 \\

~ & Ours & \textbf{73.26} & \textbf{63.35} & \textbf{70.11} & \textbf{107.02} & \textbf{134.07} & \textbf{85.48} & \textbf{58.83} & 80.39 &\textbf{62.52} \\

\midrule
\multirow{7}*{GPT-4V} & RS & 60.49 & 45.38 & 59.13 & 101.56 & 115.87 & 82.40 & 16.98 & 58.72 & 45.08\\

~ & I2I & 56.48 & 47.19 & 56.27& 103.41 & 110.68 & 85.17 & 18.35& 62.31 & 43.41 \\

~ & IQ2IQ & - & - &- & - & - & - & -& -& -  \\

~ & IQPR & - & - &- & - & - & - & -& -& - \\

~ & DEmO & - & - &- & - & - & - & -& -& -\\

~ & Lever-LM & \textbf{65.31} & 54.62 & 65.73 & 106.34 & 126.98 & \textbf{84.81} & 45.62 & 60.31& 48.34 \\

~ & Ours & 65.16 & \textbf{56.17} & \textbf{68.89} & \textbf{107.29} & \textbf{129.71} & 83.96 &\textbf{51.48} & \textbf{64.17} & \textbf{50.59}\\

\bottomrule

\end{tabular}}
\caption{\label{detailed}
Detailed results of different methods across all tasks for the five LVLMs used in the evaluation, with all generated sequences being 4-shot. The highest scores are highlighted in \textbf{bold}. Our model achieves the best performance in all but four tasks, demonstrating its generalization and effectiveness.
}
\end{table*}

We can go deep into the per-model results in Table \ref{detailed}. The findings are as follows: (1) TACO exhibits the best performance in all but three tasks across nine datasets and five LVLMs, demonstrating its great efficiency and generalization. Upon examining the outputs, we observe that GPT-4V tends to deviate from the ICD format and produce redundant information more easily than open-source LVLMs, aligning with \citep{wu2023gpt4v}. This results in the quality improvement of the ICL sequence not always translating into stable ICL performance gains for GPT-4V, which may explain why TACO did not achieve the best performance in two of its tasks. (2) For tasks like VizWiz and \textbf{Hybrid}, TACO consistently improves the quality of sequence generation in all LVLMs compared to similarity-based models, demonstrating the importance of increasing task semantics for complex task mappings. We find that the performance gains from TACO are not directly related to the model's intrinsic ability on these tasks. Unlike simpler tasks like classification, for tasks with complex mappings, task semantics still has a significant impact, even when LVLMs exhibit strong few-shot learning abilities. This shows that models with strong ICL capabilities on certain tasks retain, and even strengthen, their ability to leverage task semantics, underscoring the value of improving ICL sequence quality.

From all the above experiments, we can conclude that TACO effectively constructs prompts to maintain a coherent global task mapping. In this mapping, latent task signals from each demonstration are effectively integrated. Consequently, LVLMs can extract and synthesize fine-grained, task-specific information. This indicates that the key to superior performance lies in the prompt’s ability to align with the underlying task intent, enabling deeper reasoning and more accurate outputs. The results in Table 2 and the corresponding analysis in Section 5 further explain TACO's good performance. The decline in performance observed when these components are removed indicates that maintaining a cohesive global mapping from individual demonstrations is fundamental to enabling the model to leverage task-relevant features during inference. Moreover, dynamic encoding during the ICL sequence configuration helps preserve and optimize the task mapping in the autoregressive process, thereby enhancing prompt quality.

\textbf{Efficiency analyses.} ICL is widely adopted for its efficiency \cite{rwkv}; therefore, efficiency was a primary focus in the design of TACO. TACO exhibits high efficiency during both training and inference. Firstly, TACO is a lightweight language model composed solely of a fusion module and four transformer decoder blocks. With only 140M parameters, its training cost is extremely low compared to LLMs. Moreover, owing to its specialized training objectives, TACO prioritizes data quality over sheer quantity, enabling effective training with a relatively small amount of high-quality data. Simultaneously, our approach to constructing training data is highly efficient. By leveraging LVLM for self-assessment, we significantly reduce the overhead associated with incorporating external metrics. For instance, when using CIDEr to construct training data for the image captioning task, the costs are nearly 9 times higher than those of our current method. To further validate TACO's training efficiency, we compare its training time with that of 4-layer Lever-LM on the same training sets.
\begin{table*}[t]
  \centering
  \resizebox{\textwidth}{!}{%
    \begin{tabular}{lccccccccc}
      \toprule
      \textbf{Method}    & VQAv2 & VizWiz & OK-VQA & Flickr30K & MSCOCO & HatefulMemes & \textbf{Hybrid} & \textbf{Fast}  & \textbf{CLEVR}  \\
      \midrule
      Lever-LM  &  9.96 &   4.72 &   2.95 &      5.13 &   5.56 &         2.67 &   6.37 &  2.08 &   1.54 \\
      TACO    & 10.33 &   4.69 &   3.04 &      5.21 &   5.53 &         2.85 &   6.46 &  1.92 &   1.41 \\
      \bottomrule
    \end{tabular}%
  }
\caption{GPU hours (h) consumed during training by two models.}
\label{app:training}
\end{table*}
Table \ref{app:training} demonstrates that TACO achieves superior performance compared to the baseline while maintaining comparable training costs, and even requires less training time than the baseline on several datasets. This evidence substantiates the high training efficiency of TACO. Moreover, to test inference efficiency, we compare different methods' retrieval time—that is, the time required to construct a 4-shot ICL sequence from instances in a specific dataset for a given query sample.
\begin{table*}[t]
  \centering
    \resizebox{\textwidth}{!}{%
    \begin{tabular}{*{10}{c}}
      \toprule
      \textbf{Method}    & VQAv2 & VizWiz & OK-VQA & Flickr30K & MSCOCO & HatefulMemes & \textbf{Hybrid} & \textbf{Fast}  & \textbf{CLEVR} \\
      \midrule
      RS        & 0.367 & 0.209  & 0.195  & 0.271     & 0.348  & 0.187        & 0.361  & 0.142 & 0.083 \\
      IQPR      & 0.639 & 0.301  & 0.287  & 0.574     & 0.725  & 0.352        & 0.701  & 0.291 & 0.200 \\
      Lever-LM  & 0.392 & 0.234  & 0.204  & 0.293     & 0.354  & 0.195        & 0.383  & 0.149 & 0.089 \\
      TACO    & 0.387 & 0.227  & 0.200  & 0.280     & 0.356  & 0.190        & 0.375  & 0.147 & 0.085 \\
      \bottomrule
    \end{tabular}}
    \caption{Average retrieval time (s) (4-shot) of different methods across all LVLMs.}
  \label{app:time}
\end{table*}
Table \ref{app:time} proves that TACO achieves notable performance improvements without compromising inference efficiency, with its runtime remaining comparable to that of RS.
\subsection{More VL Tasks}
\begin{table}[t]
  \centering
    \begin{tabular}{lcc}
      \toprule
      \textbf{Method}   & \textbf{GQA} & \textbf{A-OKVQA} \\
      \midrule
      RS                & 49.56        & 41.20           \\
      I2I               & 48.74       & 41.31           \\
      Lever-LM          & 56.28        & 45.93            \\
      TACO            & \textbf{57.62}& \textbf{47.80}   \\
      \bottomrule
    \end{tabular}%
  \caption{Average 4-shot results of different ICL sequence configuration methods on GQA and A-OKVQA benchmarks.}
  \label{app:more}
\end{table}
To further demonstrate the broad applicability of our method to more tasks, especially challenging VL tasks, we evaluate on two additional benchmarks: GQA \cite{gqa} and A-OKVQA \cite{aokvqa}. Both require multi-hop reasoning, offering a more rigorous assessment of the model’s performance under complex task mappings.

As shown in Table \ref{app:more}, TACO achieves the highest average results on both GQA and A-OKVQA. Consequently, our method attains optimal performance across all 11 datasets, which thoroughly demonstrates its robustness and effectiveness.

\subsection{Discussions about Oracle}
At inference time, ground truth responses are unavailable, so Oracle cannot be applied directly. Here, we adapt the pseudo-response generation idea from IQPR to Oracle. First, we generate a pseudo response using either RS or I2I; next, we treat this pseudo response as Oracle’s ground truth for greedy retrieval. This process yields two additional baseline configuration methods.
\begin{table*}[t]
  \centering
\resizebox{\textwidth}{!}{%
  \begin{tabular}{lccccccccc}
    \toprule
    Method       & VQAv2  & VizWiz & OK-VQA & Flickr30K & MSCOCO  & HatefulMemes & Hybrid  & Fast   & CLEVR   \\
    \midrule
    Oracle(RS)   & 61.17  & 42.27  & 52.98  & 95.82   & 112.71  & 75.62       & 20.72   & 66.38  & 45.73  \\
    Oracle(I2I)  & 57.93  & 41.86  & 51.82  & 97.64     & 113.93  & 75.28       & 18.39  & 65.92  & 40.96   \\
    \bottomrule
  \end{tabular}}
    \caption{Results of two Oracle-based configuration methods across 9 benchmarks.}
  \label{app:oracle}
\end{table*}
As demonstrated in Table \ref{app:oracle}, using pseudo results for Oracle can amplify the method's inherent drawbacks. On VQAv2, Oracle (I2I) performs 0.28\% lower than I2I; on VizWiz, Oracle (RS) is 1.11\% lower than RS, and Oracle (I2I) experiences an even more significant performance loss, being 1.24\% lower than I2I. On other datasets, the performance of these two methods is also unstable. Therefore, using pseudo results to guide Oracle is a viable, yet not effective alternative. The limitations of the Oracle-based methods underline TACO’s practical utility.

\section{Additional Ablation Study}
\label{ablation}

\subsection{Input Embeddings}
\begin{table}[t]
\centering
\resizebox{\columnwidth}{!}{\begin{tabular}{lcccccc}
\hline
& VQAv2 & MSCOCO & HatefulMemes & \textbf{Hybrid} & \textbf{Fast} & \textbf{CLEVR} \\
\hline
(CLIP Encoder) &  & &  &  &  &  \\
N/A & 45.38 & 111.57 & 70.21 & 37.67 & 60.84 & 43.52 \\
Adapter only & 47.29 & 112.42 & 73.26 & 40.15 & 63.59 & 45.71 \\
Fully training & \textbf{49.63} & 115.21 & \textbf{78.49}  & \textbf{43.84} & 66.64 & \textbf{47.80} \\
Last two & 46.58 & 114.48 & 74.52 & 40.25 & 66.13 & 46.73 \\
\rowcolor[HTML]{F0F0F0}
Last three & 48.95 & \textbf{115.36} & 78.04 & 43.76 & \textbf{66.72} & 47.54 \\
\hline
(Fusion Module) &  & &  &  &  &  \\
+ Ternary fusion& 49.57 & 114.72 & 80.37 & 43.65 & 67.48 & 48.07 \\
\rowcolor[HTML]{F0F0F0}
+ Binary fusion & \textbf{52.07} & \textbf{119.47} & \textbf{80.59}& \textbf{45.22} & \textbf{68.73} & \textbf{48.45} \\
\hline
\end{tabular}}
\caption{\label{emb} Results of TACO with different input embedding configurations. (CLIP Encoder) section shows the results without adding fusion modules under various training methods for CLIP encoders. N/A indicates no training or modification. (Fusion Module) section presents the results with two fusion modules added on top of the encoders trained with the method of training the last three layers.}
\end{table}

To investigate the impact of input embedding construction on ICL sequence configuration, we vary both the training method of the CLIP encoders and the adoption of the fusion module to evaluate TACO's performance under different settings. For the CLIP encoders, we explore three alternative methods: one involves freezing its parameters and adding an MLP adapter to its output, which is then trained; another involves fully training the entire encoder; and the third involves training only the last two layers. For constructing the embeddings of multimodal ICD tokens, we first experiment with direct concatenation without fusion modules:
\begin{equation}
e_{i}=E_{I}(I_{i})+E_{T}(Q)_{i}+E_{T}(R_{i})+r_{i},
\end{equation}
where $r_{i}$ is a randomly initialized learnable component introduced into the embedding. Besides binary fusion, we examine a finer-grained ternary fusion module that assigns separate weights to control the contributions of all three components $I$, $Q$ and $R$:
\begin{equation}
   e_{i}=f_{I}\cdot E_{I}(I_{i}) +f_{Q}\cdot E_{T}(Q_{i})+f_{R}\cdot E_{T}(R_{i}),
\end{equation}
where $f_{I}, f_{Q}$ and $f_{R}$ denote the weights computed using a softmax function applied to the linear transformations, ensuring their sum equals 1. Additionally, we apply regularization to the weights: $f_{I}^2+f_{Q}^2+f_{R}^2\le \theta$ to prevent excessive reliance on specific components. 

The training approach for CLIP affects the feature representation of embeddings, which in turn influences TACO's ability to capture cross-modal details during sequence configuration. From Table \ref{emb} we observe that for tasks with intrinsic features like VQA and \textbf{Hybrid}, leaving the CLIP unchanged or only adding an adapter leads to significant degradation in the quality of the ICL sequence generation. In fact, even methods that only train the last two layers show a more noticeable performance gap compared to the current approach. This highlights that the output pattern of the third-to-last layer of the encoder is crucial for capturing core task features in multimodal ICD. When we replaced our current training method with one that fully trains CLIP, we did not observe a significant performance drop. This suggests that TACO's treatment of ICDs as tokens does not cause feature loss. In contrast, through task-aware attention, it enhances feature representation, helping mitigate the limitations of the embedding itself. Considering the high cost of training the entire encoders, current method is optimal.

As we point out in §\ref{MI}, it is important for the model to focus on fine-grained features within the two modalities for multimodal ICL. However, Table \ref{emb} shows that the use of a ternary fusion mechanism to obtain more refined embeddings actually results in worse performance compared to binary fusion, likely due to insufficient parameter capacity in TACO.
\subsection{Instruction}
\label{ipp}
Sections \ref{2.2} and \ref{shot} highlight the importance of $Inst$ in improving multimodal ICL performance. However, as shown in Table \ref{Insttable}, using the original embedding of $Inst$ to initialize $TG$ degrades TACO performance due to semantic redundancy from long text embeddings, which can cause $TG$ deviation and hinder convergence.

We further examine how the style and relative position of the $Inst$ affect performance. Its placement within the prompt is particularly critical: in standard and most effective ICL settings, all ICDs are positioned immediately before the query sample. Consequently, varying the relative position of $Inst$ serves as a direct probe of positional effects within the ICL sequence. Two new styles are developed and placed at the beginning of the prompt, while the CoT-style $Inst$ is also tested between the ICDs and query sample, as well as at the end. Diverse prompt samples are provided in Appendix \ref{instapp}. Table \ref{Insttable} shows that, although the position of $Inst$ in the prompt has only a minor overall effect on performance, placing $Inst$ at the beginning yields the greatest relative gain. However, its style significantly affects performance, with the CoT-style being the most effective. Meanwhile, results in Table \ref{style} indicate that when the instruction used for $TG$ initialization and the one included in the prompt have different styles, TACO demonstrates greater robustness. Changes in the style of $Inst'$ not only result in minimal performance degradation but also lead to significantly smaller performance variations. In contrast, for LVLMs, changes in $Inst$ style cause noticeable performance gaps and a clear preference for specific styles. This indicates that the performance fluctuations caused by $Inst$ are primarily attributable to LVLMs rather than TACO itself. Thus, $Inst$ can be viewed as a special ICD, contributing high-level local task mapping that integrates into the LVLM's global task mapping.
\begin{table}[t]
\centering
\resizebox{\columnwidth}{!}{\begin{tabular}{lcccccc}
\hline
\textbf{Instruction} & VizWiz & MSCOCO & HatefulMemes & \textbf{Hybrid} & \textbf{Fast} & \textbf{CLEVR} \\
\hline
\rowcolor[HTML]{F0F0F0}
\textit{Beginning1} & \textbf{52.07} & 119.47& 80.59 & \textbf{45.22} & \textbf{68.73} &  \textbf{48.45}\\
$Inst'\to Inst$ & 46.21 & 114.36 & 78.23 & 39.64 & 63.52 & 43.70 \\
\hline
\textit{Beginning2} & 49.61 & 118.78 & 79.13 & 44.15 & 67.42 & 46.38 \\
\textit{Beginning3} & 49.25 & 118.23 & 78.49 & 43.71 & 67.29 & 45.62 \\
\hline
\textit{Middle} & 51.85 & 119.62 & \textbf{80.6}2 & 45.18 & 68.53 & 48.27 \\
\textit{End} & 51.73 & \textbf{119.67} & 80.37 & 44.96 & 68.59 & 47.89 \\
\hline
\end{tabular}}
\caption{\label{Insttable} Results of TACO with diverse instruction types. The highest scores are highlighted in \textbf{bold}. $Inst' \to Inst$ means using $Inst$ during the initialization of $TG$.}
\end{table}
\begin{table}[t]
    \centering
    \resizebox{\columnwidth}{!}{\begin{tabular}{c|c|cccc}
    \hline
    $Inst'$ & $Inst$ & VQAv2 & VizWiz & OK-VQA & \textbf{Hybrid} \\
    \hline
    \multirow{3}{*}{\textit{Beginning1}} & \textit{Beginning2} & 62.24& 49.18&58.77 & 42.53\\
    \cline{2-2}
    & \textit{Beginning3} &61.37 & 49.26& 57.30& 42.07\\
    \cline{2-2}
    & \textit{End} & 63.58&50.69 & 59.62&42.26 \\
    \hline
    \textit{Beginning2} & \multirow{3}{*}{\textit{Beginning1}} & 65.28& 51.05& 60.81& 44.61\\
    \cline{1-1}
    \textit{Beginning3} & & 64.62& 51.28& 59.14& 44.28\\
    \cline{1-1}
    \textit{End} & &65.40 &49.08 & 60.47& 43.72\\
    \hline
    \end{tabular}}
    \caption{Results of TACO under various $Inst'$-$Inst$ combinations. $Inst'$ represents the style used for initializing $TG$, while $Inst$ refers to the style actually incorporated into the prompt.}
    \label{style}
\end{table}

\subsection{Generalization Test}
To demonstrate the generalization of TACO beyond image-to-text tasks, we evaluate its performance on NLP and text-to-image tasks. We first use the latest LLM ICL benchmark, ICLEval's \cite{chen2024icl} Rule Learning part to construct a mixed-task NLP dataset and test it on Qwen2-7B and LLaMA3-8B. For text-to-image tasks, we use the Fast Counting dataset from the VL-ICL Bench and test it on Emu2-Gen \cite{sun2024emu1}.  The ICDs in both tasks can be represented as $(Q, R)$. Results in Table \ref{gen} show that TACO consistently outperforms baselines across all tasks, highlighting its strong generalizability and wide application potential.
\begin{table}[t]
\centering
\resizebox{0.9\columnwidth}{!}{\begin{tabular}{lcc|c}
\hline
\multirow{2}*{\textbf{Methods}} & \multicolumn{2}{c|}{\textbf{NLP}}& \textbf{text-to-image} \\
\cmidrule(lr){2-4}
 & Qwen2-7B & LLaMA3-8B & Emu2-Gen\\
\hline
RS & 0.29 & 0.30& 43.67\\
Q2Q & 0.48 & 0.54&  47.83 \\
QPR & 0.47& 0.56 & 49.06 \\
Lever-LM & 0.50 & 0.60 & -\\
\rowcolor[HTML]{F0F0F0}
Ours & \textbf{0.52} & \textbf{0.61} & \textbf{51.18} \\
\hline
\end{tabular}}
\caption{\label{gen} Results of different ICL sequence configuration methods in NLP and text-to-image tasks. Both training and generated shots are set to 4. The highest scores are highlighted in \textbf{bold}.}
\end{table}
\begin{figure}[t]
  \centering
  \includegraphics[width=\columnwidth]{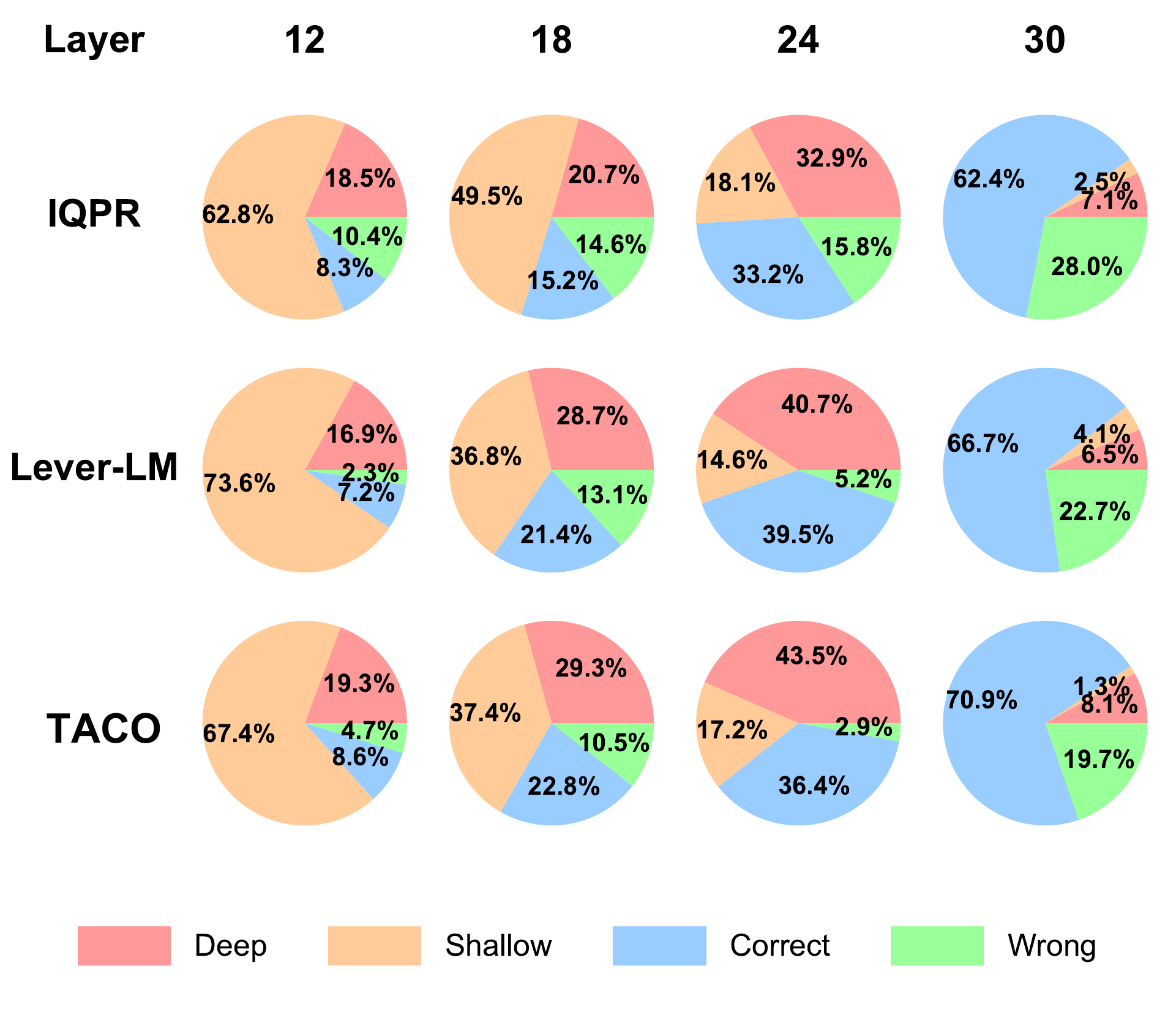}
  \caption{Visualization of the in-context lens for different methods under the Harder-Mapping setting.}
  \label{fig:inlens2}
\end{figure}
For NLP evaluation, we utilize the Rule Learning part of the latest benchmark, ICLEval. ICLEval is designed to assess the ICL abilities of LLMs, focusing on two main sub-abilities: exact copying and rule learning. The Rule Learning part evaluates how well LLMs can derive and apply rules from examples in the context. This includes tasks such as format learning, where models must replicate and adapt formats from given examples, and order and statistics-based rule learning, where the model must discern and implement patterns such as item sequencing or handling duplications. These tasks challenge LLMs to go beyond language fluency, testing their ability to generalize from context in diverse scenarios. Examples of $(Q,R)$ pairs can be found in Table \ref{icleval}. For all tasks, we use exact match scores to evaluate the predictions against the labels.

For text-to-image evaluation, we utilize the Fast Counting task in the VL-ICL Bench. In this task, artificial names are associated with the counts of objects in the image. The task is to generate an image that shows a given object in quantity associated with the keyword (e.g. perpo dogs where perpo means two). Thus, each $Q$ is a two-word phrase such as "perpo dogs", and its corresponding $R$ is an image of two dogs.

The ICDs in both tasks can be represented as $(Q, R)$. In NLP, both $Q$ and $R$ are text; in text-to-image, $Q$ is text while $R$ is an image. We simply need to adjust the embedding encoder and fusion module accordingly. The baselines are RS, \textbf{Q2Q} (Query-to-query), \textbf{QPR} (Query\&pseudo-response), and Lever-LM (not applicable to text-to-image).
\begin{figure}[ht]
  \centering
  \includegraphics[width=\columnwidth]{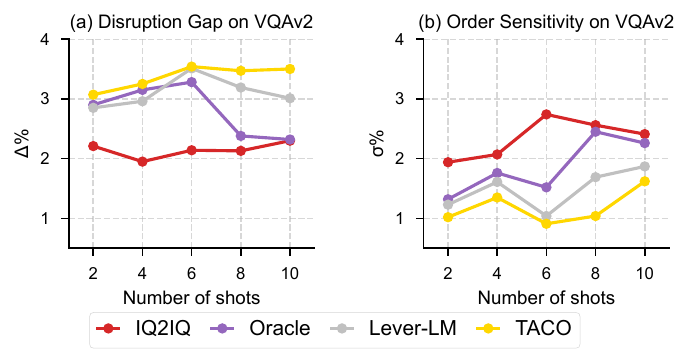}
  \caption{Analysis of task mapping cohesion in $n$-shot ICL sequences generated by different methods.}
  \label{seq}
\end{figure}
\begin{table}[ht]
  \centering
  \resizebox{\columnwidth}{!}{%
  \begin{tabular}{lcc}
    \toprule
    \textbf{Method}   & \textbf{HatefulMemes(Standard)} & \textbf{HatefulMemes(HM)} \\
    \midrule
    IQPR              & 73.62                           & 68.31                     \\
    Lever-LM          & 78.86                           & 72.15                     \\
    TACO            & \textbf{80.59}                  & \textbf{74.87}            \\
    \bottomrule
  \end{tabular}}
  \caption{Comparison of different methods under a Harder Mapping setting.}
  \label{app:harder}
\end{table}

\begin{table*}[t]
\centering
\begin{tabular}{m{3cm}<{\centering} | m{6cm}| m{5cm}}
\toprule
\textbf{Task} & \textbf{$Q$} &\textbf{$R$}\\
\midrule
Format rules & |Index|name|age|city|\newline
|---|---|---|---|\newline
|1|Elijah Morgan|36|Pittsburgh| & <person>\newline
<name>Elijah Morgan</name>\newline <age>36</age> \newline<city>Pittsburgh</city>\newline
</person>\\

\midrule
Statistics rules & 588 and 823 are friends.\newline
885 and 823 are friends. \newline
795 and 588 are friends.\newline
890 and 823 are friends.\newline
885 and 588 are friends.\newline
890 and 588 are friends.\newline
795 and 823 are friends.\newline
Query: Who are the friends of 885? & 823, 588 \\

\midrule
Order rules& Input: activity, brief, wonder, anger\newline Output: anger, wonder, activity, brief \newline Input: market, forever, will, curve \newline Output: curve, will, market, forever \newline Input: pain, leading, drag, shoot \newline Output: shoot, drag, pain, leading \newline Input: shopping, drama, care, start \newline  Output: &start, care, shopping, drama\\

\midrule
List Mapping & Input: [1, 3, 6, 1, 83]\newline
Output: [3]\newline
Input: [5, 6, 35, 3, 67, 41, 27, 82]\newline Output: [6, 35, 3, 67, 41]\newline
Input: [8, 45, 6, 18, 94, 0, 1, 2, 7, 34]\newline Output: [45, 6, 18, 94, 0, 1, 2, 7]\newline Input: [2, 7, 66, 6, 93, 4, 47]\newline
Output: & [7, 66] \\
\bottomrule
\end{tabular}
\caption{\label{icleval} The examples of four Rule Learning tasks in ICLEval. }
\vspace{-15pt}
\end{table*}

\subsection{Revisiting Task Mapping Framework}
In this section, we apply TACO’s experimental results to the task mapping theoretical framework outlined in §\ref{MI}, thereby further validating its effectiveness and generality.

We first utilize the two metrics introduced in §\ref{bas}, Disruption Gap (\textbf{$\Delta$}) and Order Sensitivity (\textbf{$\sigma$}), to evaluate task mapping cohesion in ICL sequences generated by TACO. Figure \ref{seq} shows that TACO achieves the highest $\Delta$ and lowest $\sigma$ across all shots. This not only indicates that TACO-generated ICL sequences construct robust task mappings effectively utilized by LVLMs but also provides further evidence supporting the validity of our task mapping framework. Notably, from the results at shots 8 and 10, we observe that although TACO's training data is constructed by Oracle, it overcomes the cohesion weakening caused by bias accumulation through task mapping augmentation.

Next, we employ HatefulMemes’s Harder-Mapping setting to evaluate TACO’s performance on more challenging specific-mapping tasks. Results in Table \ref{app:harder} indicate that shifting task mapping from standard to harder reduces the performance of all three methods, but TACO still achieves the highest scores under both settings. Harder-Mapping setting increases the difficulty of understanding task mapping in ICDs, preventing the model from bypassing deep reasoning through parametric knowledge. In contrast, TACO guides LVLMs to generate ICL sequences with clearer, more identifiable global mappings, enabling them to overcome the comprehension barriers introduced by more challenging mappings.

Besides evaluating performance, we also employ in-context lens to examine the average evolution of the internal outputs in LVLM by these methods under HM settings. Figure \ref{fig:inlens2} illustrates the evolution of the LVLM’s internal reasoning as it learns from ICL sequences generated by different methods. The results show that sequences produced by TACO enable the model to most effectively infer task mappings and generate accurate responses.
\end{document}